\documentclass[journal]{IEEEtran}
\usepackage{tikz}
\usepackage{pgfplots}
\usetikzlibrary{shapes,arrows,positioning}
\definecolor{RBRed}{RGB}{215,0,18}
\definecolor{RBLightred}{RGB}{234,104,118}
\definecolor{RBMagenta}{RGB}{168,1,99}
\definecolor{RBLightmagenta}{RGB}{208,103,173}
\definecolor{RBViolet}{RGB}{63,19,108}
\definecolor{RBLightviolet}{RGB}{150,124,177}
\definecolor{RBDarkblue}{RGB}{8,66,126}
\definecolor{RBLightdarkblue}{RGB}{109,154,188}
\definecolor{RBLightblue}{RGB}{14,120,197}
\definecolor{RBLightlightblue}{RGB}{111,185,226}
\definecolor{RBTurquoise}{RGB}{19,153,160}
\definecolor{RBLightturquoise}{RGB}{111,201,204}
\definecolor{RBLightgreen}{RGB}{103,180,25}
\definecolor{RBLightlightgreen}{RGB}{174,219,125}
\definecolor{RBDarkgreen}{RGB}{10,81,57}
\definecolor{RBLightdarkgreen}{RGB}{110,162,147}
\definecolor{RBDarkgray}{RGB}{66,76,88}
\definecolor{RBLightdarkgray}{RGB}{153,159,166}
\definecolor{RBLightgray}{RGB}{178,179,181}
\definecolor{RBLightlightgray}{RGB}{215,215,218}

\ifCLASSINFOpdf
  \usepackage{graphicx}	
  \graphicspath{{./images/}}
  \DeclareGraphicsExtensions{.pdf,.jpeg,.png}
\else
\fi
\usepackage{amsmath,amssymb}
\DeclareMathOperator{\E}{\mathbb{E}}
\ifCLASSOPTIONcompsoc
  \usepackage[caption=false,font=normalsize,labelfont=sf,textfont=sf]{subfig}
\else
  \usepackage[caption=false,font=footnotesize]{subfig}
\fi
\usepackage{fixltx2e}

\usepackage{stfloats}
\usepackage{tabularx}

\usepackage{hyperref}

\usepackage{booktabs}

\usepackage{makecell}

\usepackage[scrtime]{prelim2e}
\renewcommand{\PrelimWords}{This work has been submitted to the IEEE for possible publication. Copyright may be transferred without notice, after which this version may no longer be accessible.}

\usepackage{changes}

\newcommand{\TTC}{\text{TTC}}
\newcommand{\ROI}{\text{ROI}}
\newcommand{\TPR}{\text{TPR}}
\newcommand{\FPR}{\text{FPR}}
\newcommand{\TP}{\text{TP}}
\newcommand{\FN}{\text{FN}}
\newcommand{\FP}{\text{FP}}
\newcommand{\TN}{\text{TN}}

\newcommand{\unit}[1]{\ensuremath{\, \mathrm{#1}}}

\newcommand{\metric}[3]{%
	\begin{tabular}[c]{@{}r@{}} {\scriptsize \color{gray} \raisebox{-0.2mm}{$\mathsf{#3}$}} \\ $#1$ \\ {\scriptsize  \color{gray} \raisebox{0.7mm}{$\mathsf{#2}$}} \end{tabular}%
}%

\begin{document}

%
\title{Pedestrian Behavior Prediction\\for Automated Driving:\\Requirements, Metrics, and Relevant Features}
%
%
%

\author{Michael~Herman, J\"{o}rg~Wagner, Vishnu~Prabhakaran, Nicolas~M\"{o}ser,\\Hanna~Ziesche, Waleed~Ahmed, Lutz~B\"{u}rkle, Ernst~Kloppenburg, and~Claudius~Gl\"{a}ser%
\thanks{M.~Herman, J.~Wagner, V.~Prabhakaran, H.~Ziesche, and E.~Kloppenburg are with the Bosch Center for Artificial Intelligence, Germany.}
\thanks{N.~M\"{o}ser, L.~B\"{u}rkle, and C.~Gl\"{a}ser are with the Robert Bosch GmbH, Corporate Research, Germany.}
\thanks{W.~Ahmed is with the Robert Bosch GmbH, Cross-Domain Computing Solutions -- Automated Driving, Germany.}
\thanks{e-mail: Michael.Herman@de.bosch.com}}

%
%

\markboth{Manuscript submitted for review to IEEE Transactions on Intelligent Transportation Systems}%
{tbd \MakeLowercase{\textit{et al.}}: Pedestrian Behavior Prediction for Automated Driving: Requirements, Metrics, and Relevant Features}
%



\maketitle

\begin{abstract}
Automated vehicles require a comprehensive understanding of traffic situations to ensure safe and anticipatory driving. In this context, the prediction of pedestrians is particularly challenging as pedestrian behavior can be influenced by multiple factors. In this paper, we thoroughly analyze the requirements on pedestrian behavior prediction for automated driving via a system-level approach. To this end we investigate real-world pedestrian-vehicle interactions with human drivers. Based on human driving behavior we then derive appropriate reaction patterns of an automated vehicle and determine requirements for the prediction of pedestrians. This includes a novel metric tailored to measure prediction performance from a system-level perspective. The proposed metric is evaluated on a large-scale dataset comprising thousands of real-world pedestrian-vehicle interactions. We furthermore conduct an ablation study to evaluate the importance of different contextual cues and compare these results to ones obtained using established performance metrics for pedestrian prediction. Our results highlight the importance of a system-level approach to pedestrian behavior prediction.
\end{abstract}


%
\IEEEpeerreviewmaketitle


\section{Introduction}
%
%
%
%

\IEEEPARstart{R}{oad} safety is a key driver for the development of Driver Assistance (DA) and Automated Driving (AD) systems. According to a report of the World Health Organization \cite{WorldHealthOrganization2018} traffic accidents cause more than 1.3 million fatalities annually, almost half of them being Vulnerable Road Users (VRUs). Therefore, the protection of VRUs, in particular pedestrians, constitutes 
a major goal of intelligent vehicles. The Automatic Emergency Braking system for Pedestrians (AEB-P) is a good example on how driver assistance systems already protect pedestrians today. AEB-P detects pedestrians in the predicted vehicle's path and, if a collision cannot be avoided by the driver, automatically initiates emergency breaking. By either avoiding the collision or, if avoidance is not possible, reducing the velocity of an impact, pedestrian AEB systems mitigate pedestrian fatality and injury \cite{Haus2019}. The detection of pedestrians and the prediction of their behavior are essential components of an AEB-P system. Prediction of an AEB-P is generally restricted to short prediction horizons in the order of 1 to $2 \unit{s}$ and is typically based on kinematic models.

On the other hand, automated driving not only addresses near-collision situations, but broadens the scope to everyday driving scenarios. Thus, besides collision mitigation, comfortable driving that imitates human driving behavior shifts into focus. In order to react appropriately at an early stage automated driving requires an extended pedestrian behavior prediction to correctly reason about a situation on a longer time-scale. 

\begin{figure}[!t]
	\centering
	\includegraphics[width=1.0\linewidth]{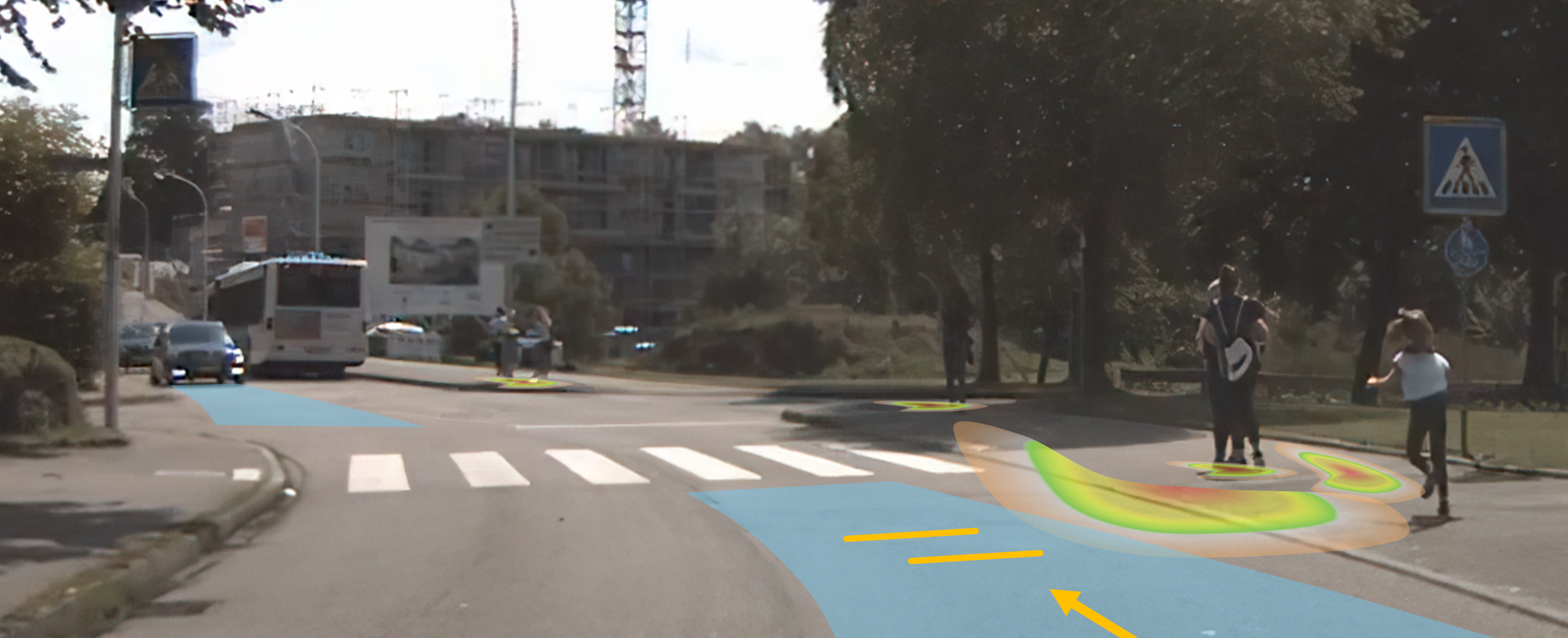}
	\caption{Exemplary scenario in which the behavior of a pedestrian depends on multiple contextual cues, e.g. the present road infrastructure or interactions with other traffic participants.}
	\label{fig:pedPredTeaser}
\end{figure}

Recent years have seen an increased interest in using deep learning based methods for the purpose of long-term pedestrian prediction. Most of the works on this topic however used generic metrics both for development and evaluation of the prediction models. While these generic metrics are suitable for measuring the overall accuracy of a predicted behavior, they do not take into account the actual requirements of downstream functions, like e.g. an automated driving system. We argue that due to this, important task-specific requirements are not considered in model development and evaluation, since generic metrics do not or only partially cover those requirements. As a result the proposed models are often too complex or suboptimal for the downstream task. Therefore, we propose a new function-specific metric, which is based on system-level requirements. This metric allows for a more task-informed assessment of models for pedestrian prediction.

Especially on a longer time-scale the behavior of pedestrians cannot be investigated in isolation, but rather has to be considered within the context of the overall traffic scene. The exemplary scenario of a girl running along a sidewalk depicted in Fig.~\ref{fig:pedPredTeaser} illustrates this relationship: The future behavior of the girl is likely to depend on a variety of factors. This includes the static driving infrastructure (e.g. the road layout, the zebra crossing), interactions of the pedestrian with other traffic participants (e.g. an approaching vehicle), and appearance or communication cues (e.g. gestures). Yet, the importance and influence of individual contextual cues on prediction accuracy and on the downstream AD function is not obvious a priori.

In this paper, we derive requirements and a performance metric for pedestrian behavior prediction in the context of automated driving. For evaluation purposes, we present a prediction model based on a Conditional Variational Autoencoder (CVAE) that specifically addresses long-term prediction by including contextual cues of the traffic scene. Finally, we investigate the importance of contextual cues in terms of prediction accuracy. Our results highlight the importance of a system-level approach to pedestrian behavior prediction. In detail, the contributions of the paper are:
\begin{itemize}
	\item An appropriate system reaction pattern for interactions of an automated vehicle with pedestrians is derived from an analysis of human driving behavior.
	\item Requirements for pedestrian behavior prediction in automated driving are specified and a corresponding metric to assess prediction performance is derived.
	\item On the basis of the proposed metric, a CVAE prediction model is evaluated on a large-scale dataset comprising thousands of real-world pedestrian-vehicle interactions.
	\item The relevance of different contextual cues is assessed based on an ablation study and the results are compared to the ones obtained using established performance metrics for pedestrian prediction.
\end{itemize}

The remainder of this paper is organized as follows: We first give an overview on related work in Sec.~\ref{sec:relatedWork}. We then introduce our large-scale dataset of vehicle-pedestrian interactions in Sec.~\ref{sec:dataset}. In Sec.~\ref{sec:requirements}, we analyze pedestrian-vehicle interactions to determine human driving behavior and to derive requirements on pedestrian prediction for automated driving. Our prediction model is introduced in Sec.~\ref{sec:rnnModel} and thoroughly evaluated in Sec.~\ref{sec:results}. Finally, we conclude the paper with a discussion of our results in Sec.~\ref{sec:discussion}.


\section{Related Work}
\label{sec:relatedWork}

In this section, we give an overview of state of the art approaches as well as typical input features applied to pedestrian prediction. Furthermore, evaluation metrics and public datasets are briefly summarized.

\subsection{Pedestrian Prediction Approaches}

Pedestrian prediction has already been studied for a long time, resulting in numerous approaches that address this problem in the automotive domain \cite{Ridel2018, Camara2020} and beyond \cite{Rudenko2020}. It is important to note, however, that in the context of Advanced Driver Assistance Systems (ADAS) and AD the employed output representation of prediction models and consequently the system integration may significantly differ. Approaches range from the prediction of pedestrian crossing intentions \cite{SchneemannHeinemann2016, Voelz2016}, over walking destinations \cite{Rehder2015, Rehder2018}, to the prediction of paths \cite{deo2021trajectory} or trajectories \cite{radwan2018multimodal}, where the latter can further be distinguished into trajectory prediction in image-view and bird's-eye view. 

We consider a prediction of bird's-eye view pedestrian trajectories best suited for an AD system as the downstream planning of appropriate system reactions usually relies on this kind of representation. That is why we will focus on such approaches in the following.

To the best of our knowledge, requirements on pedestrian behavior prediction for AD have not been thoroughly analyzed so far. We strongly believe that it is essential to mirror pedestrian prediction to requirements of the downstream task in order to obtain an optimal overall system performance. By taking a system-level approach to pedestrian behavior prediction, this paper provides an important contribution for this.

\subsection{Trajectory Prediction Models}

Many traditional approaches for predicting the future motion of traffic participants depend on a set of explicitly defined dynamics equations that are generally derived from physics-based motion models \cite{Rudenko2020}. Often, these approaches are used in combination with Probabilistic Graphical Models \cite{mazor1998interacting, gindele2010probabilistic, agamennoni2012estimation}.

Recently, pattern-based methods that learn behavioral patterns from data have outperformed traditional approaches. Especially, deep learning based solutions became state of the art for most of the problems related to public datasets. Often, Recurrent Neural Networks (RNN) are used for encoding trajectories of interacting agents and decoding future behavior \cite{alahi2016social, deo2018multi}. One of the major problems associated with these approaches is to accurately capture the probabilistic, multi-modal distribution over trajectories. In order to address this issue recent deep learning based methods predict parametric distributions \cite{alahi2016social}, learn mixtures of Gaussian trajectory distributions \cite{chai2020multipath}, use adversarial training approaches \cite{kuefler2017imitating}, or introduce discrete \cite{tang2019multiple, salzmann2020trajectron} or continuous \cite{lee2017desire, felsen2018will, bhattacharyya2019conditional, casas2020implicit} latent variables.

For the investigations in this paper, we use models that build on Conditional Variational Autoencoders (CVAE) \cite{sohn2015learning}. Their use of continuous latent variables renders them suitable for capturing complex, multi-modal probability distributions.

\subsection{Contextual Cues}
Human behavior is influenced by contextual cues of internal and external stimuli. The survey \cite{Rudenko2020} groups them into three categories:
cues of the target agent, the dynamic environment, and the static environment. Potential target agent cues are the motion state (e.g. position, velocities) \cite{agamennoni2012estimation, alahi2016social, deo2018multi}, appearance-based cues (e.g. head or body pose) \cite{Kooij2014}, or semantic attributes (e.g. age or gender) \cite{Ma2017}. While traditional models often do not take into account influences of the dynamic environment \cite{Elnagar1998, Bennewitz2005, Scholler2020CVM}, other approaches exist that model interactions with other agents \cite{Trautman2010,Kuderer2012,alahi2016social} or even social groups \cite{Robicquet2016}. Regarding cues of the static environment, there are several approaches that neglect this influence \cite{Schneider2013,Luber2012}, some others only model influence of individual static objects \cite{Bartoli2018}, while still others model more complex influences from environment geometry and topology \cite{tang2019multiple,Gao2020}. In addition, recent work \cite{Rasouli2019PedestrianActionAnticipation} studies the contribution of different contextual cues on an action classification task.

\subsection{Evaluation Metrics} \label{sec:eval-metrics}

Performance evaluation is an integral part of the process of creating
a prediction model. The survey \cite{Rudenko2020} extensively
discusses different metrics for models that predict
trajectories. These metrics fall into two classes, geometric and
probabilistic. A widely used geometric metric is the
Average Displacement Error (ADE), which applies to models providing
point predictions. ADE measures the euclidean distance of a predicted
trajectory from ground truth positions at a specific prediction time interval,
averaged over the trajectory, or over multiple trajectories. Quite commonly 
the ADE metric is also applied to probabilistic predictions by averaging
over the predictive distribution as well.

Probabilistic metrics are used for models that provide predictions in the form of probability densities.  This kind of metrics measures how well the predictions capture the uncertainties inherent to the prediction process as well as the true process. A typical metric here is average Negative Log Likelihood (NLL) of the ground truth positions. Unfortunately, these kind of metrics tend to lack intuitive interpretability. One subtlety with the different probabilistic metrics is whether they encourage multimodal predictive distributions or not.

This raises the question whether the metrics discussed so far measure properties relevant for possible applications of the model under consideration. The authors of \cite{Theis2015d} discuss the problem of evaluating generative (probabilistic) models and come to the conclusion that application specific metrics are generally required.

\subsection{Existing Datasets}

\begin{table*}[ht!]
	\centering
	\caption{Comparison of Existing Datasets}
	\begin{tabular}{lccp{1pt}cccp{1pt}cc}
		\toprule
		{\bf Dataset} & \multicolumn{2}{c}{\bf Representation} && \multicolumn{3}{c}{\bf Features} && \multicolumn{2}{c}{\bf Size} \\
		              & \parbox{3em}{\centering image plane} & \parbox{3em}{\centering top view} && map & ego & \makecell{pedestrian\\ attributes} && \makecell{recording time\\ \mbox{[min]}} & \makecell{\# pedestrian tracks\\ (with attributes / total)} \\
		\toprule
		STIP \cite{liu2020spatiotemporal}           & x &   &&   &   &   &&    923 &      - / 25,000 \\
		JAAD \cite{Rasouli2017}                     & x &   &&   &   & x &&     43 &    686 /  2,786 \\
		PIE \cite{Rasouli2019}                      & x &   &&   & x & x &&    360 &  1,842 /  1,842 \\
		TITAN \cite{malla2020titan}                 & x &   &&   & x & x &&    175 &  8,592 /  8,592 \\
		\midrule
		SDD \cite{Robicquet2016}                    &   & x && x &   &   &&    620 &      - / 11,216 \\
		INTERACTION \cite{interactiondataset}       &   & x && x &   &   &&    991 &      - /  1,700 \\
		inD \cite{inDdataset}                       &   & x && x &   &   &&    600 &      - / 3,107 \\
		\midrule
		Argoverse \cite{Argoverse}                  & x & x && x & x &   &&     30 &      - /  1,322 \\
		nuScenes \cite{nuscenes2019}                & x & x && x & x &   &&    333 &      - / 11,512 \\
		\midrule
		PePScenes \cite{Rasouli2020}                & x & x && x & x & x &&    333 &    719 / 11,512 \\
		ours                                        & x & x && x & x & x &&  4,434 &  9,438 / 93,162 \\
		\bottomrule
	\end{tabular}
	\label{tab:relatedwork_datasets}
\end{table*}

In contrast to the large number of datasets devoted to pedestrian detection there is only a limited number of datasets available that address pedestrian prediction in an automated driving context. Table~\ref{tab:relatedwork_datasets} summarizes and compares the most important ones. The comparison takes into account the employed representation (image plane vs. top view), the availability of features which are relevant for pedestrian prediction (a semantic map, ego-vehicle data, and detailed pedestrian attributes at a perceptual and/or behavioral level), and the size of the datasets (in terms of the number of unique pedestrian tracks and recording time).

Datasets from the first group provide video recordings including annotated objects. Some of them additionally include rich behavioral annotations for pedestrians. The most notable contributions to this group include the Joint Attention in Autonomous Driving (JAAD) dataset \cite{Rasouli2017}, the Pedestrian Intention Estimation (PIE) dataset \cite{Rasouli2019}, and the TITAN dataset \cite{malla2020titan}. However, even though these datasets are suitable for benchmarking vision-based algorithms for pedestrian detection or intention recognition, they are limited in use for investigating trajectory prediction where an object representation in bird's-eye view is required.

Datasets from the second group like the Stanford Drone Dataset (SDD) \cite{Robicquet2016}, the INTERACTION dataset \cite{interactiondataset}, or the Intersection Drone dataset (inD) \cite{inDdataset} comprise trajectories from a bird's-eye view and are thus suitable for pedestrian prediction. However, since drones were used to record these datasets they are limited to a small number of locations, and more important, do not provide detailed pedestrian attributes.

Recently, various large scale automotive datasets were published e.g. Argoverse \cite{Chang_2019_CVPR} and nuScenes \cite{Caesar_2020_CVPR}, many of which also comprise tracking or motion forecasting challenges. However, none of these datasets provide detailed pedestrian attributes and mainly focus on the prediction of vehicles trajectories. Furthermore, they lack a significant number of scenes with pedestrian-vehicle interactions. The authors of \cite{Rasouli2020} conducted a post-labeling of nuScenes to provide additional pedestrian attributes. However, for their PePScenes dataset only 6\% of the pedestrians in nuScenes were taken into account, since all others were not of interest for the driving task.


\section{Dataset}
\label{sec:dataset}


\begin{figure}[b!]
	\centering
	\subfloat[round courses]{\includegraphics[width=0.48\linewidth]{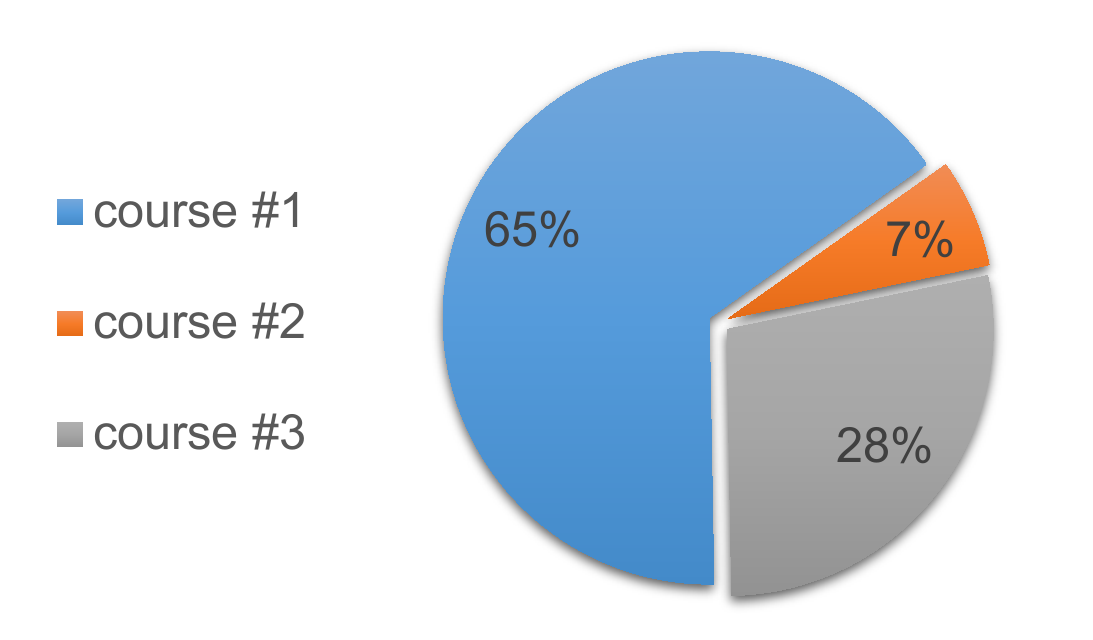}
		\label{fig:dataset_statistics_course}}
	\hfil
	\subfloat[crossing location]{\includegraphics[width=0.48\linewidth]{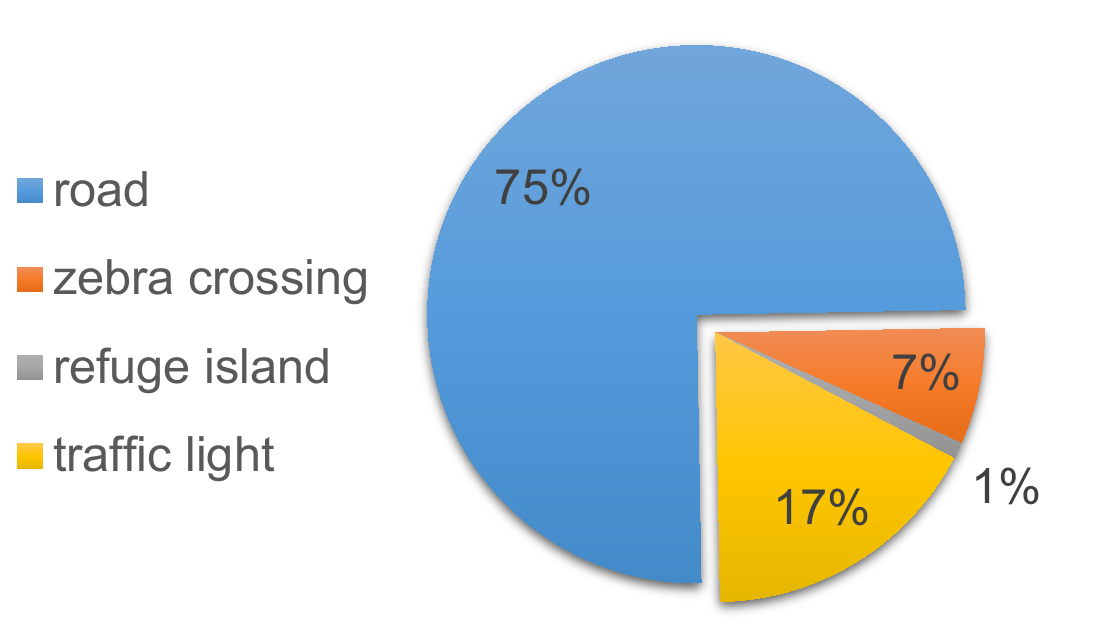}
		\label{fig:dataset_statistics_crossing}}
	\caption{Dataset statistics: (a) distribution of pedestrian tracks with respect to the three courses; (b) locations at which pedestrians crossed the street.}
	\label{fig:dataset_statistics}
\end{figure}

To overcome the limitations of existing datasets, we created a large-scale dataset that specifically addresses pedestrian prediction in automated driving context. In detail, the requirements on the datasets were as follows:
\begin{itemize}
	\item 2D pedestrian positions in the ground plane. 
	\item Annotations for additional features that are relevant for the prediction task. Most notably, a semantic map of the road infrastructure, ego-vehicle data for modeling pedestrian-vehicle interactions, and additional appearance-based attributes for pedestrians.
	\item A large number of pedestrians that are relevant for the driving task. The respective pedestrian-vehicle interactions shall cover various scenarios, e.g. interactions at different traffic control elements.
\end{itemize}

\subsection{Data Collection}

The dataset we refer to in this paper comprises vehicle-pedestrian interactions from inner-city traffic in southern Germany. The data was recorded on three different round-courses with lengths between 2 and 4~km. The routes were chosen to maximize variability of traffic scenarios including both downtown and suburban areas, different road sizes, traffic densities, and number of pedestrians. Furthermore, the routes contain various traffic control elements which are relevant for vehicle-pedestrian interactions, most notably zebra crossings, pedestrian refuge islands, or a combination of both.

To further increase scenario coverage a number of actors were positioned at different locations along the courses. The instruction of actors followed a semi-scripted approach where actors were told not to perform specific interactions with the recording vehicle but rather to arbitrarily vary their walking routes, interactions and behavior in a realistic manner. Since actors had to adapt their behavior to the respective traffic situation, including the recording vehicle and other traffic participants, the setup proved to result in a great variety of realistic vehicle-pedestrian interactions.

The recordings were carried out during three weeks in fall of 2018. Overall, we recorded 74~hours of data using seven different drivers to reduce driving style biases. Data acquisition was performed using a Bosch test vehicle equipped with various surround sensors. The recorded data comprises 3D point clouds of a 360 degree LiDAR sensor and images of two front-facing cameras with horizontal opening angles of 45 and 90 degrees, respectively. Furthermore, an IMU with differential GPS provides precise information on the ego-vehicle's position and motion. 


\subsection{Data Labeling}

\begin{figure}[!t]
	\centering
	\includegraphics[width=1.0\linewidth]{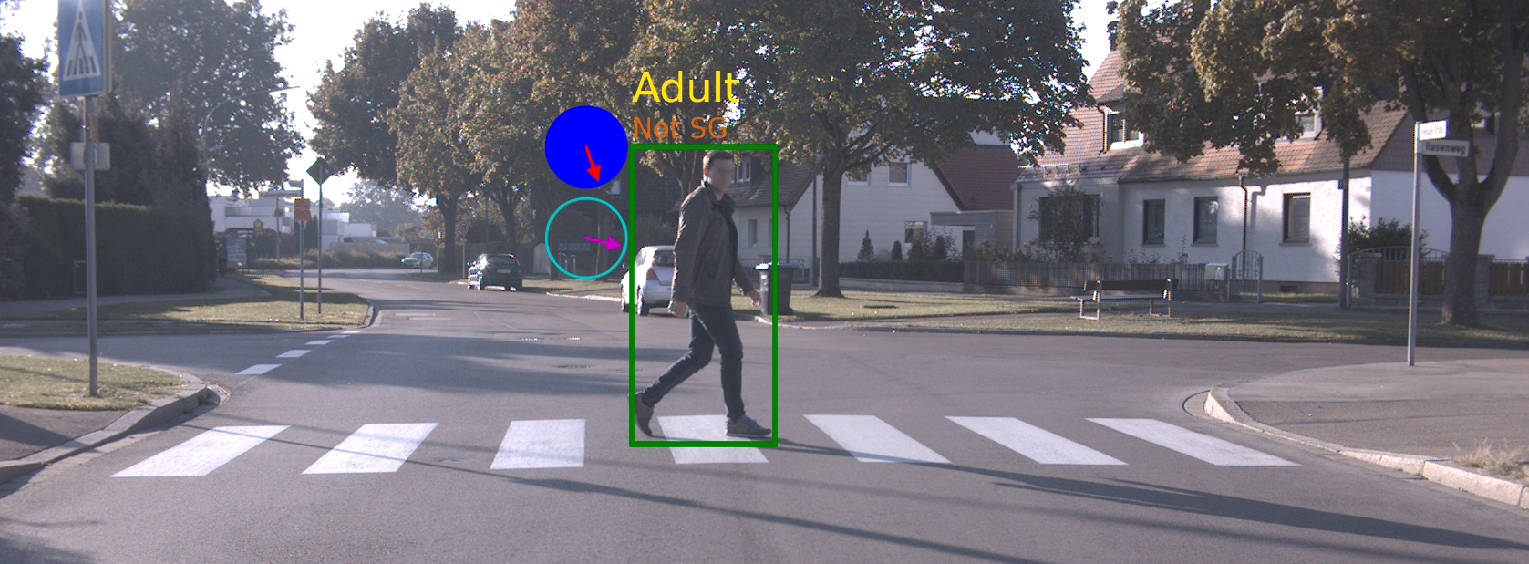}
	\caption{Example of labeled pedestrian attributes: The compass plots depict head and body orientation, where the filled upper compass indicates that the pedestrian is looking at the ego-vehicle.}
	\label{fig:dataset_pedFeatureLabels}
\end{figure}

\begin{figure}[!b]
	\centering
	\includegraphics[width=1.0\linewidth]{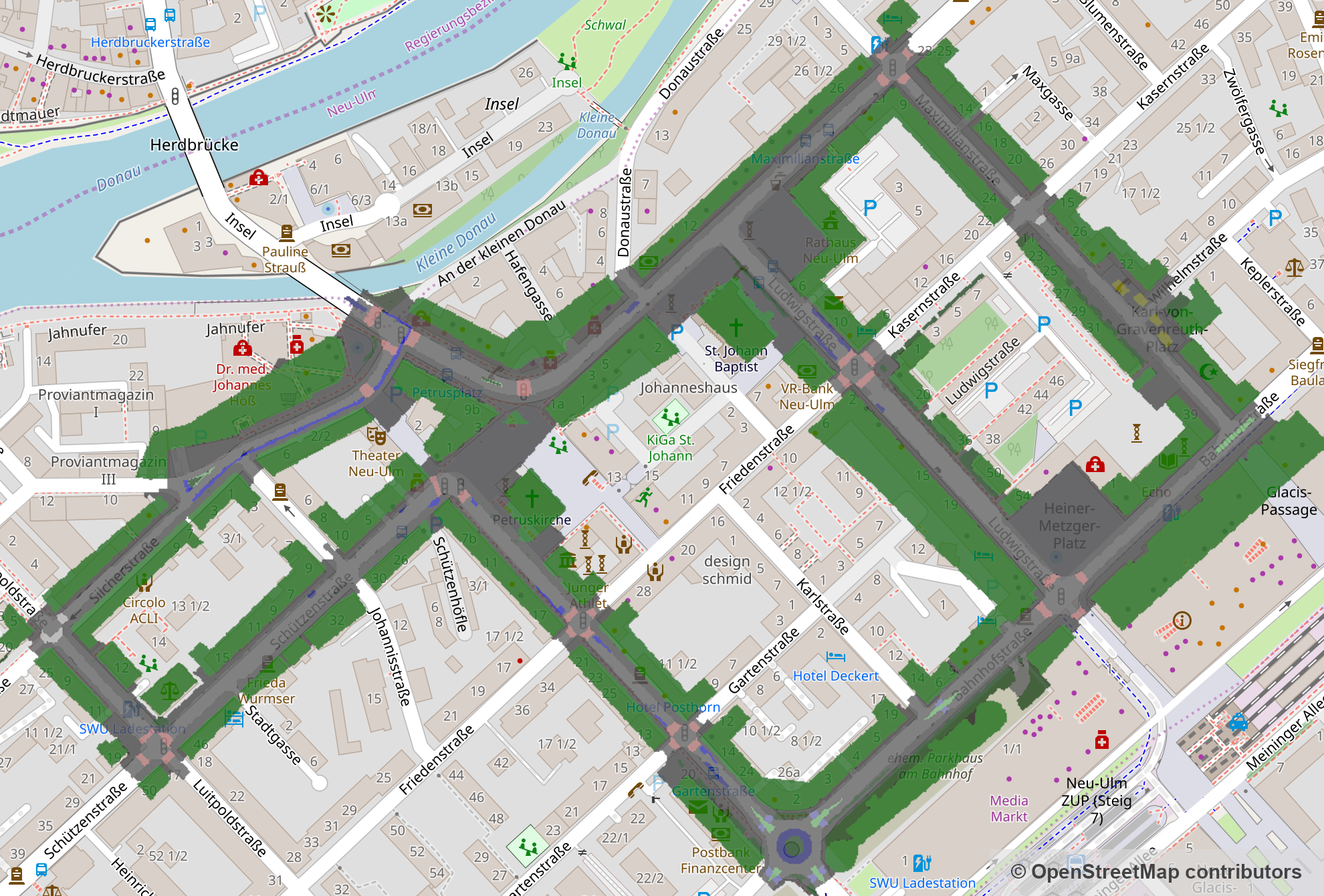}
	\caption{Semantic map shown on top of a road map. Colors denote different semantic classes. {$\copyright$ \href{https://www.openstreetmap.org/copyright}{OpenStreetMap} contributors (\href{https://openstreetmap.org}{openstreetmap.org}, \href{https://opendatacommons.org}{opendatacommons.org})}}
	\label{fig:dataset_roundCourseAndMap}
\end{figure}

Data post processing included an automatic extraction of pedestrian trajectories. As a first step, we applied a Mask\-RCNN object detector to the recorded image data. Pedestrian detections were subsequently lifted to 3D by matching them with respective LiDAR point clusters. To this end, LiDAR point clouds were ego-motion compensated, projected onto the image plane, and finally matched based on their overlap with pedestrian masks. The resulting 3D detections were tracked by Kalman filtering using a constant velocity model and a greedy association scheme. Finally, the 3D annotations were projected into the ground plane to obtain 2D pedestrian positions.

Overall, 93,162 unique pedestrian tracks with an average track length of 6.6\unit{s} were extracted from the recorded data, including approximately 5,500 actor trajectories. 7~\% of all pedestrian trajectories are crossing the road, out of which approximately 40~\% originate from actors. 
This demonstrates that even though actor trajectories constitute only a small portion of the overall dataset, it was possible to significantly increase scenario coverage by employing actors during the recordings. 
Fig.~\ref{fig:dataset_statistics} illustrates the distribution of pedestrian tracks with respect to the different round courses and crossing locations.
For our further investigations we excluded pedestrian tracks at traffic lights because behavior there is primarily determined by the traffic light states.

To enable an investigation of potentially behavior-relevant features, a subset of 9,438 pedestrian tracks was manually labeled. For the labeling, the tracks were randomly selected while ensuring that tracks are balanced between crossing and non-crossing pedestrians as well as location (i.e. presence of different traffic control elements).  As depicted in Fig.~\ref{fig:dataset_pedFeatureLabels}, the frame-wise labels include body and head orientation (in degrees).

Finally, we created semantic maps of the round courses via hand-labeling of decimeter-level accurate aerial orthoimages (see Fig.~\ref{fig:dataset_roundCourseAndMap}). These maps comprise the locations of roads, sidewalks, cycle tracks, bus lanes, barred areas, zebra crossings, refuge islands, traffic lights, lawn, and buildings. The trajectories of the ego-vehicle as well as those of the detected pedestrians hence can be projected onto the map, given the precise global position recordings of the ego-vehicle.

The large variety of scenarios and rich feature description renders the dataset suitable for pedestrian behavior prediction. However, the recordings only comprise German traffic. Thus, the generalization of trained prediction models and our results may be limited to regions with similar cultural factors or traffic rules.


\section{Requirements on Pedestrian Behavior Prediction}
\label{sec:requirements}


\subsection{Analysis of Human Driving Behavior}

\begin{figure*}[hb]
	\centering
	\subfloat{\includegraphics[width=0.45\textwidth]{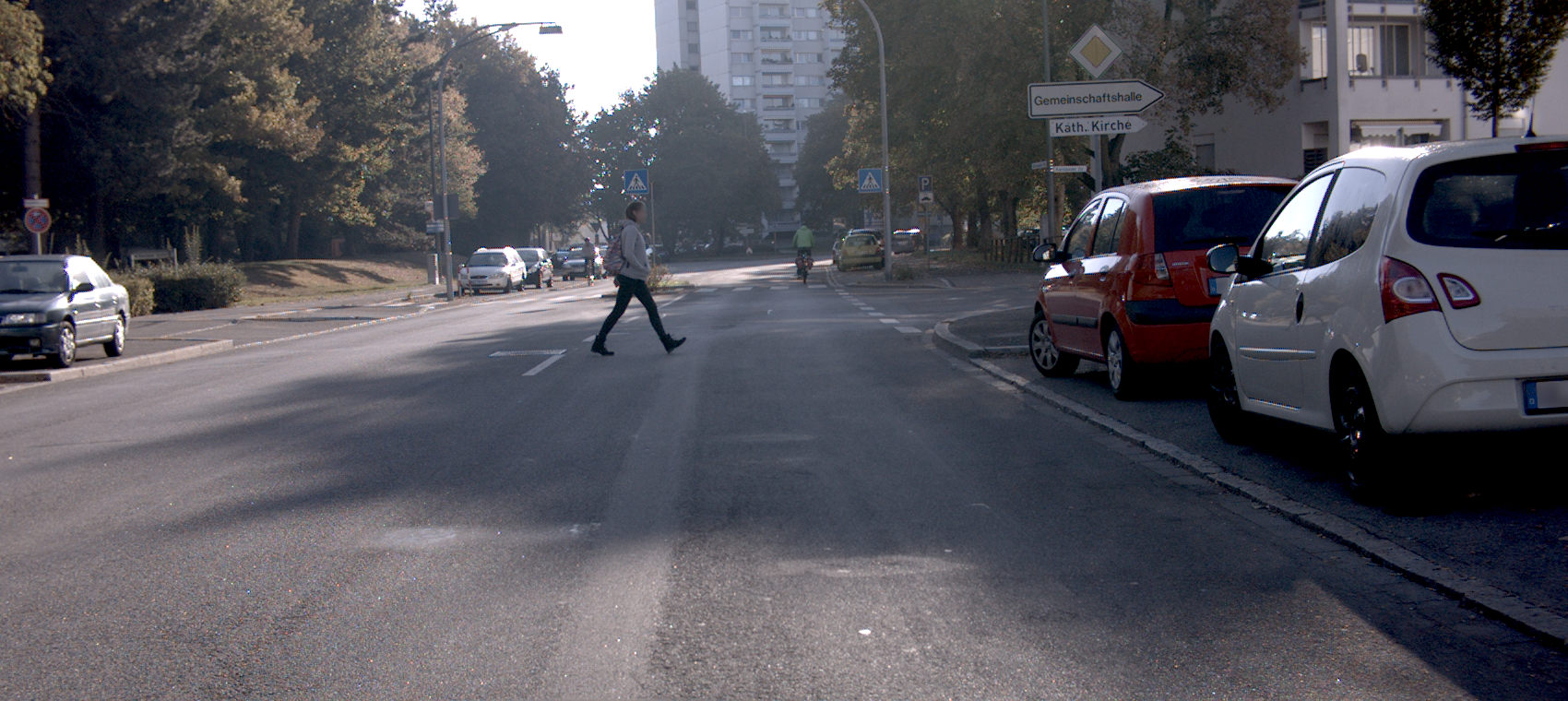}}
	\hfil
	\subfloat{\includegraphics[width=0.45\textwidth]{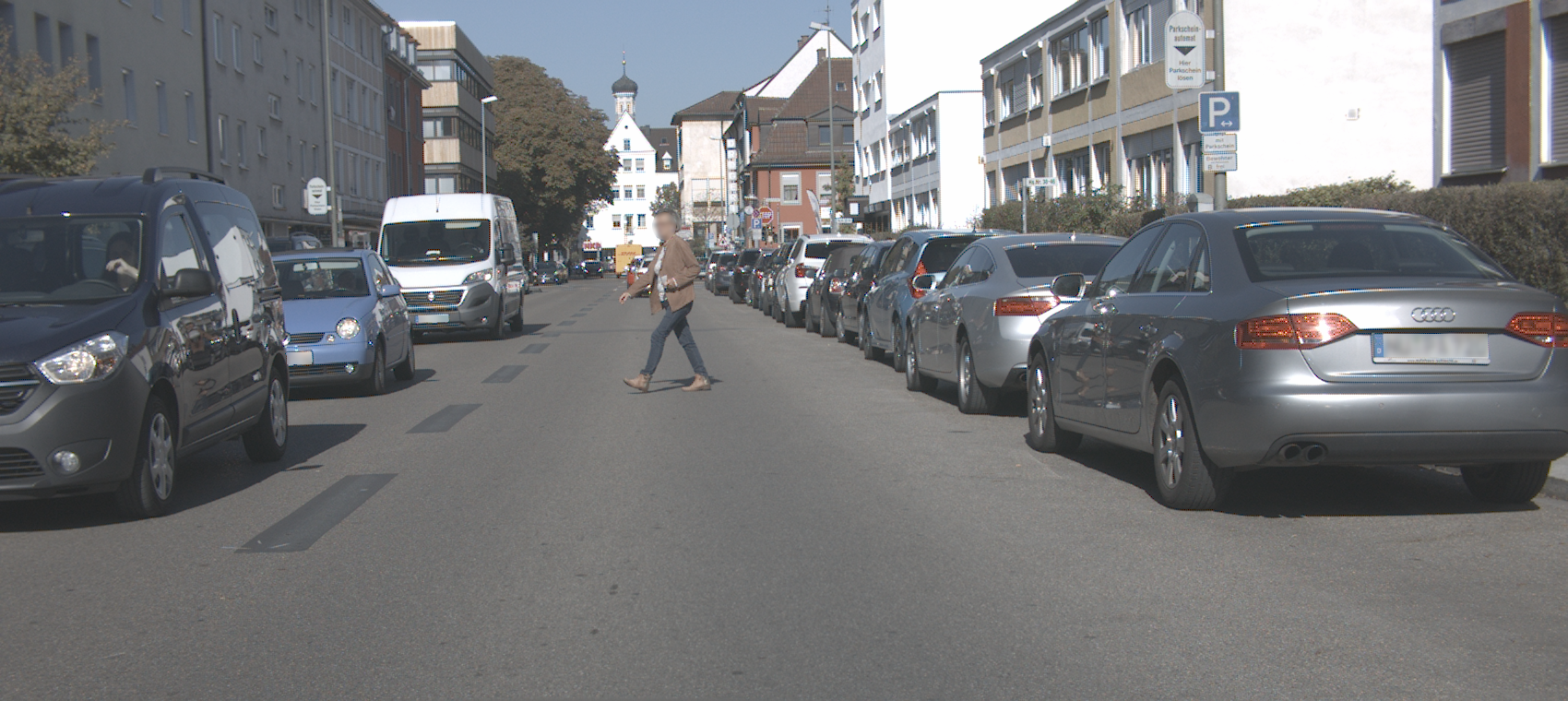}}
	\vskip 0pt
	\setcounter{subfigure}{0}
	\subfloat[]{\includegraphics[width=0.475\textwidth]{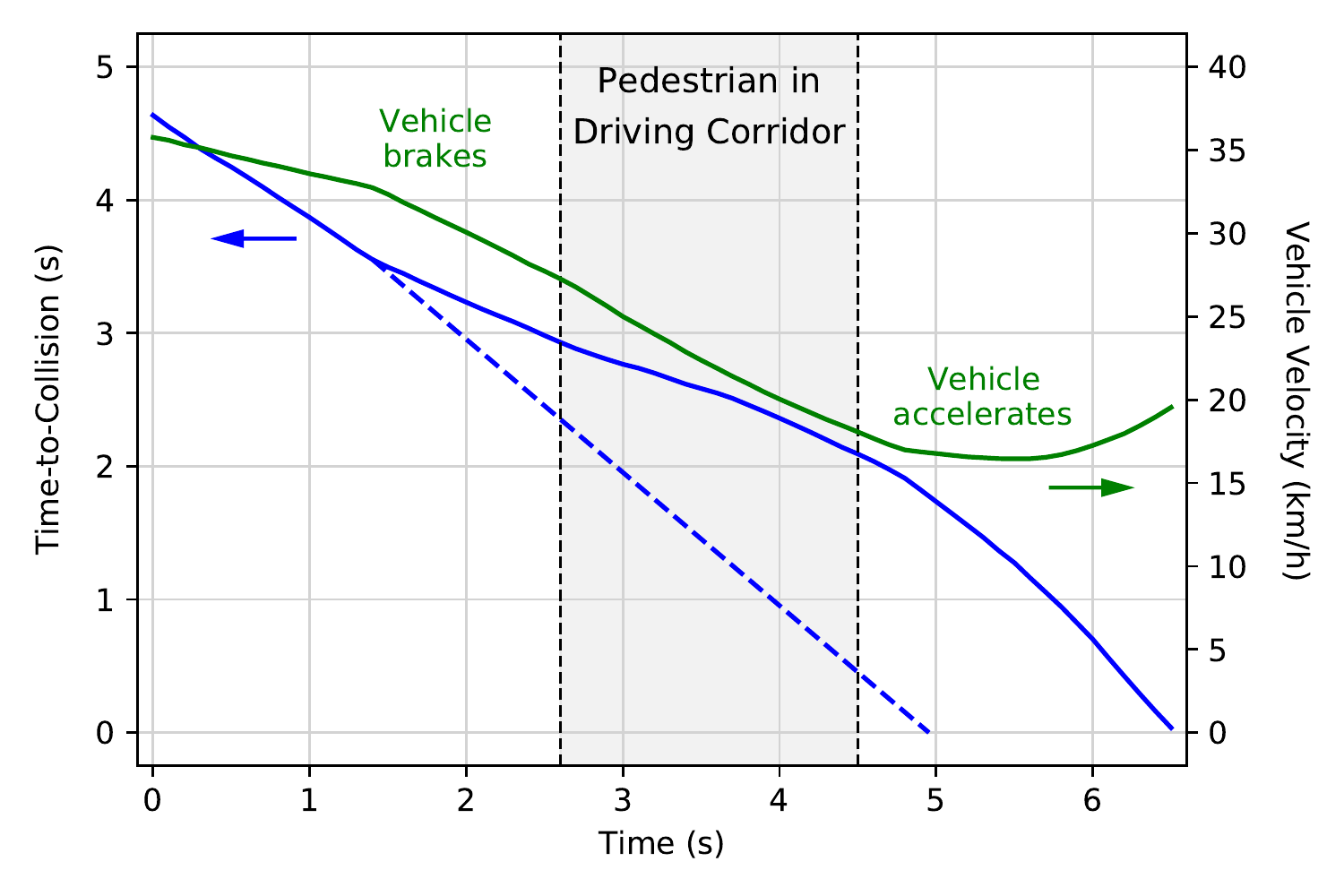}
		\label{fig:req_humanDriverAnalysis_a}}
	\hfil
	\subfloat[]{\includegraphics[width=0.475\textwidth]{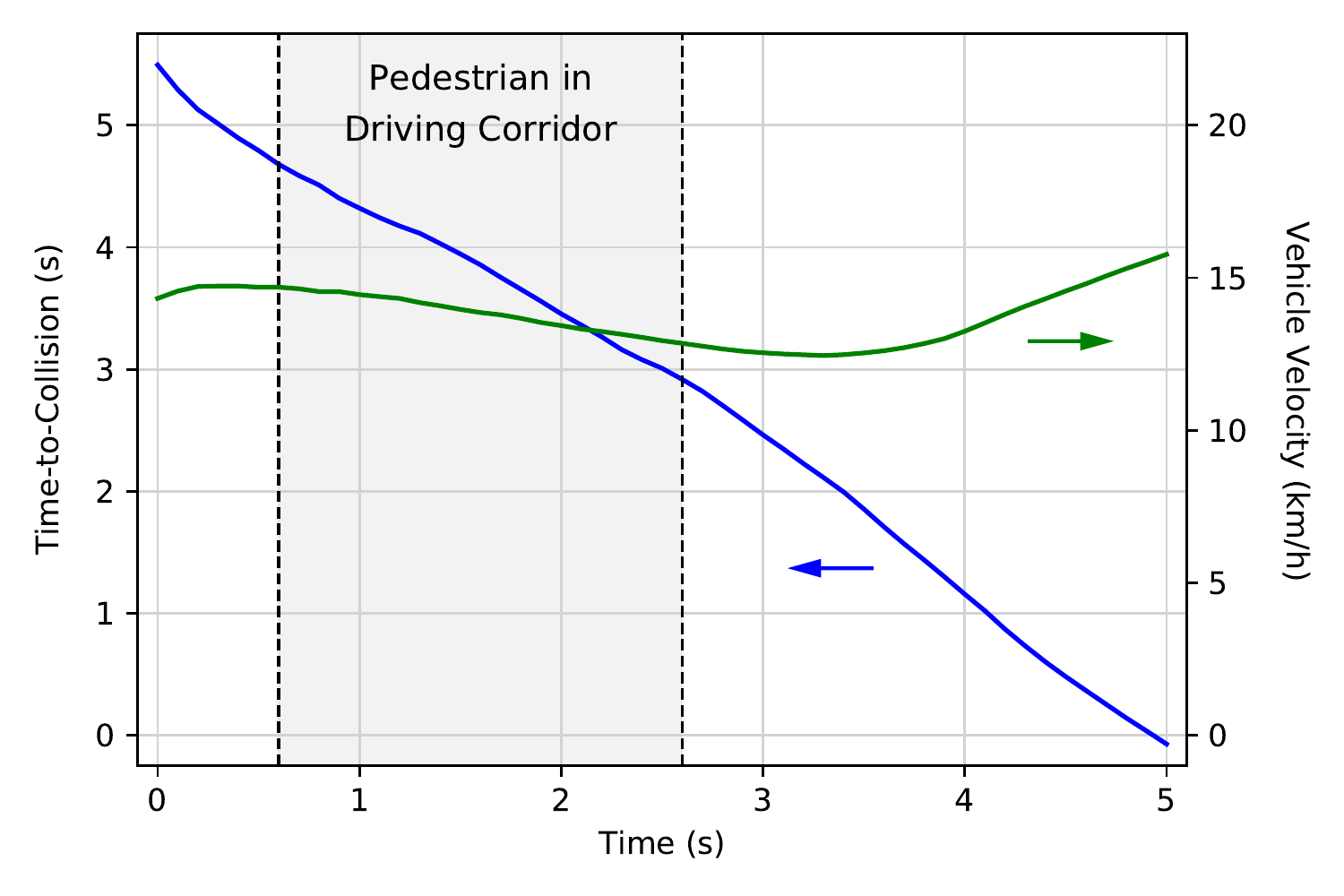}
		\label{fig:req_humanDriverAnalysis_b}}	
	\hfil
	\caption{Two traffic scenarios of a vehicle approaching a pedestrian crossing the roadway. In scenario (a) the driver brakes in order to maintain a Time-to-Collision ($\TTC$) above approximately $2 \unit{s}$ while the pedestrian is traversing the driving corridor of the vehicle. In scenario (b) the driver maintains a sufficiently large time gap between his vehicle and the pedestrian without having to brake.}
	\label{fig:req_humanDriverAnalysis}
\end{figure*}

The interaction of vehicles and pedestrians has been studied extensively in recent years \cite{Rasouli2020Survey}. While early studies on the crossing behavior of pedestrians date back to the 1950s, the complex interaction process between drivers and pedestrians that e.g. occurs at crosswalks is a relatively new field of research \cite{SchneemannGohl2016}. At roadways without areas distinctly labeled for pedestrian crossing, it is the pedestrian's responsibility to find a safe gap in traffic. Generally, the minimum gap which is still accepted by pedestrians when they decide to cross the road in front of an approaching vehicle is denoted as acceptance gap. A common measure for how safe a specific gap is, is the Time-to-Collision $\TTC = d / v$. It is calculated from the distance $d$ of an approaching vehicle, in relation to its driving velocity $v$ and thus indicates the time required for the vehicle to arrive at the pedestrians location assuming constant velocity. Although common sense might suggest that $\TTC$ is the basis for pedestrians' gap selection it has been shown that other factors such as vehicle speed \cite{Petzoldt2014} or crossing distance \cite{Zhao2019} have an influence on the size of chosen gaps.

When interacting with pedestrians it is crucial that the system behavior of an automated vehicle is perceived neither uncomfortable nor even critical by both the pedestrian and the passengers of the vehicle. In that respect, an automated vehicle should ideally imitate the behavior of a defensive human driver \cite{Houtenbos2005}. In order to define a suitable system behavior of an automated vehicle, we investigated different scenarios where a human driver interacts with pedestrians crossing the roadway in front of the vehicle. These scenarios were taken from the dataset described in Section~\ref{sec:dataset}. Two typical examples are depicted in Fig.~\ref{fig:req_humanDriverAnalysis}.

The graph in Fig.~\ref{fig:req_humanDriverAnalysis_a} shows the velocity (green) and the $\TTC$ (blue) of a vehicle approaching a pedestrian who crosses the road from left to right. The two vertical dashed lines at $t_1= 2.6 \unit{s}$ and $t_2= 4.5 \unit{s}$ mark the points in time at which the pedestrian enters and leaves the driving corridor of the vehicle, respectively. For the analysis in this paper we assume a driving corridor width of $3 \unit{m}$. Overall it takes the pedestrian $1.9 \unit{s}$ to traverse the danger zone of the driving corridor. At time $t_1$, when they enter the corridor, the $\TTC$ of the approaching vehicle is $2.9 \unit{s}$ and at time $t_2$, when the pedestrian finally leaves the corridor, $\TTC$ is $2.1 \unit{s}$.

As soon as the driver recognizes that the pedestrian intends to cross the roadway, they slightly start to brake their vehicle at $t = 1.5 \unit{s}$. This can be clearly seen by the decrease in slope of the vehicle velocity and by the increase in slope of the $\TTC$ curves, respectively. By slightly braking, the driver maintains a $\TTC$ above approximately $2 \unit{s}$ while the pedestrian is traversing the driving corridor. As soon as the pedestrian has left the corridor and is thus out of the danger zone, the driver releases the brake at $t = 4.8 \unit{s}$ and shortly afterwards accelerates the vehicle.

In contrast, the dashed blue line in the graph indicates how the $\TTC$ would evolve over time if the driver did not brake and maintained a constant velocity. In this case, the vehicle would approach the pedestrian faster leading to smaller $\TTC$ values while the pedestrian is traversing the driving corridor. Here, the $\TTC$ would drop to a value of $0.5 \unit{s}$ at $t_2$ when the pedestrian leaves the driving corridor.

Fig.~\ref{fig:req_humanDriverAnalysis_b} depicts another traffic scenario with a vehicle approaching a pedestrian crossing the roadway from right to left. Again, the dashed vertical lines indicate when the pedestrian enters and leaves the driving corridor, respectively. In contrast to the previous case, the driver neither brakes nor accelerates while their vehicle is approaching the pedestrian. This is due to the fact that the time gap already is at a safe level $> 2.9 \unit{s}$ while the pedestrian is traversing the driving corridor and, therefore, the driver does not have to take any action but rather maintains a constant velocity of approximately $14 \unit{km/h}$.

These two examples suggest that human drivers try to establish a time gap between the pedestrian and their vehicle that does not fall below a certain threshold value for the time span while a pedestrian traverses the driving corridor. This hypothesis is also confirmed by a statistical analysis of the minimum time gaps which occur in crossing scenarios. To this end, 2,238 scenarios of pedestrians crossing the roadway from right to left and vice versa in front of an approaching vehicle were identified in the dataset. For these scenarios, we determined the minimum time gaps occurring while pedestrians were traversing the vehicle corridor.

\begin{figure}[ht]
	\centering
	\includegraphics[width=1.0\linewidth]{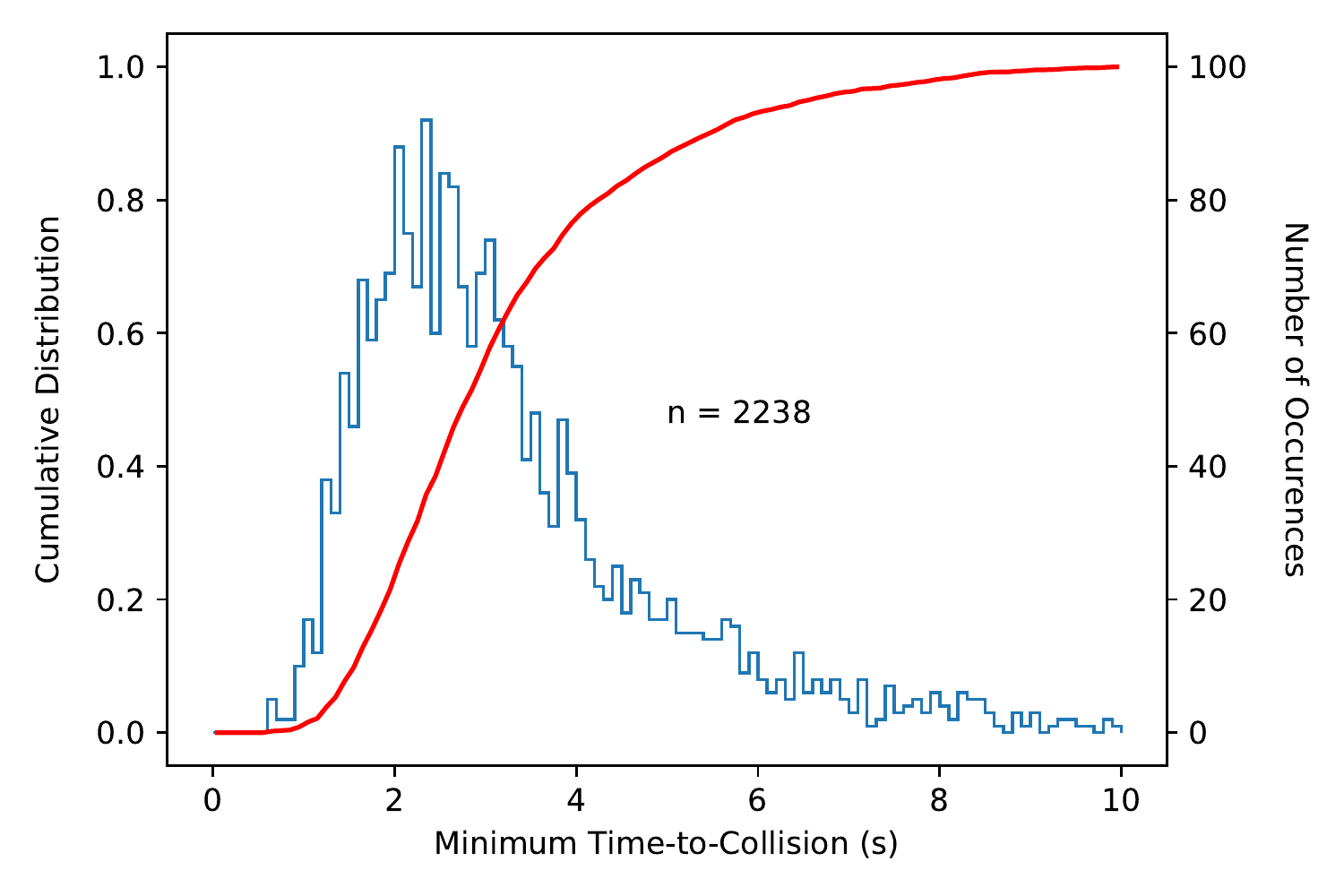}
	\caption{Cumulative distribution function (red) and number of occurrences (blue) of the minimum time gap while pedestrians traverse the driving corridor of an approaching vehicle. The median of the distribution is at a minimum time gap of $2.84 \unit{s}$.}
	\label{fig:req_humanDriverAnalysis_TTC}
\end{figure}

Fig.~\ref{fig:req_humanDriverAnalysis_TTC} shows the cumulative distribution function (red) and the number of occurrences (blue) of these scenarios as a function of the minimum time gap. The onset of individual scenarios occurs at a minimum time gap of $0.6 \unit{s}$ and the distribution reaches the maximum number of occurrences in the right-open interval between $2.3 \unit{s}$ and $2.4 \unit{s}$. The median of the distribution is at a minimum time gap of $2.84 \unit{s}$, the first and third quartiles are at $2.09 \unit{s}$ and $3.92 \unit{s}$, respectively.

The distribution function of minimum time gaps thus shows that drivers try not to fall below a certain threshold value for the time gap between their vehicle and the pedestrian crossing, which is perceived as safe and comfortable by both parties. If the initial situation of a crossing scenario leads to a time gap lower than the threshold value, the driver reacts by adjusting the speed of their vehicle accordingly, e.g. by applying the brakes. Furthermore, the result shows that a minimum time gap of at least $2.8 \unit{s}$ (i.e. the median of the distribution) is perceived as safe and comfortable by the majority of traffic participants.

As stated earlier, the acceptance gap is defined as the minimum gap still accepted by pedestrians when they decide to cross the road in front of an approaching vehicle. While it has been widely used to characterize the crossing behavior of pedestrians it is not an appropriate measure to describe the interaction between drivers and pedestrians as they cross the roadway. This is due to the fact that the acceptance gap solely depends on the pedestrians' decision to cross in a specific traffic situation and does not account for the drivers' reaction. In contrast, our results show that the minimum time gap of pedestrians while they traverse the driving corridor can be used to analyze the interaction of both parties as it combines the crossing intent of the pedestrian given a specific traffic situation and the reaction of the driver to the crossing pedestrian.

Still both metrics are related to each other: While the acceptance time gap corresponds to the $\TTC$ when a pedestrian enters the roadway, the minimum time gap usually refers to the $\TTC$ when a pedestrian is about to leave the driving corridor of the approaching vehicle. Consequently, the minimum time gap takes on smaller values than the acceptance time gap. The difference between both metrics thereby depends on the time span it takes the pedestrian to traverse the driving corridor and the reaction of the driver of the approaching vehicle.


\subsection{Derived AD System Reaction Pattern}
\label{sec:ad_system_reaction}

Based on our analysis of human driving behavior we next derive system reactions of an automated vehicle that should imitate those of human drivers. For the sake of simplicity we focus on longitudinal vehicle control, i.e. the adaption of the vehicle's velocity along a predefined or planned path. As shown in the previous section, the observed minimum time gap between a vehicle and a pedestrian traversing the driving corridor is of particular importance in this respect. Our analysis suggests, that there is a threshold value below which a situation is perceived uncomfortable or even critical. Acceptance of minimum time gaps is, of course, subjective and may depend on driving style, street layout, or traffic density. However, the histogram in Fig.~\ref{fig:req_humanDriverAnalysis_TTC} suggests that time gaps below $2 \unit{s}$ (i.e. approximately the first quartile of the distribution) are seldom observed and hence also should be avoided by an automated vehicle.

\begin{figure*}[ht]
	\centering
	\subfloat[without system reaction]{\includegraphics[width=0.45\textwidth]{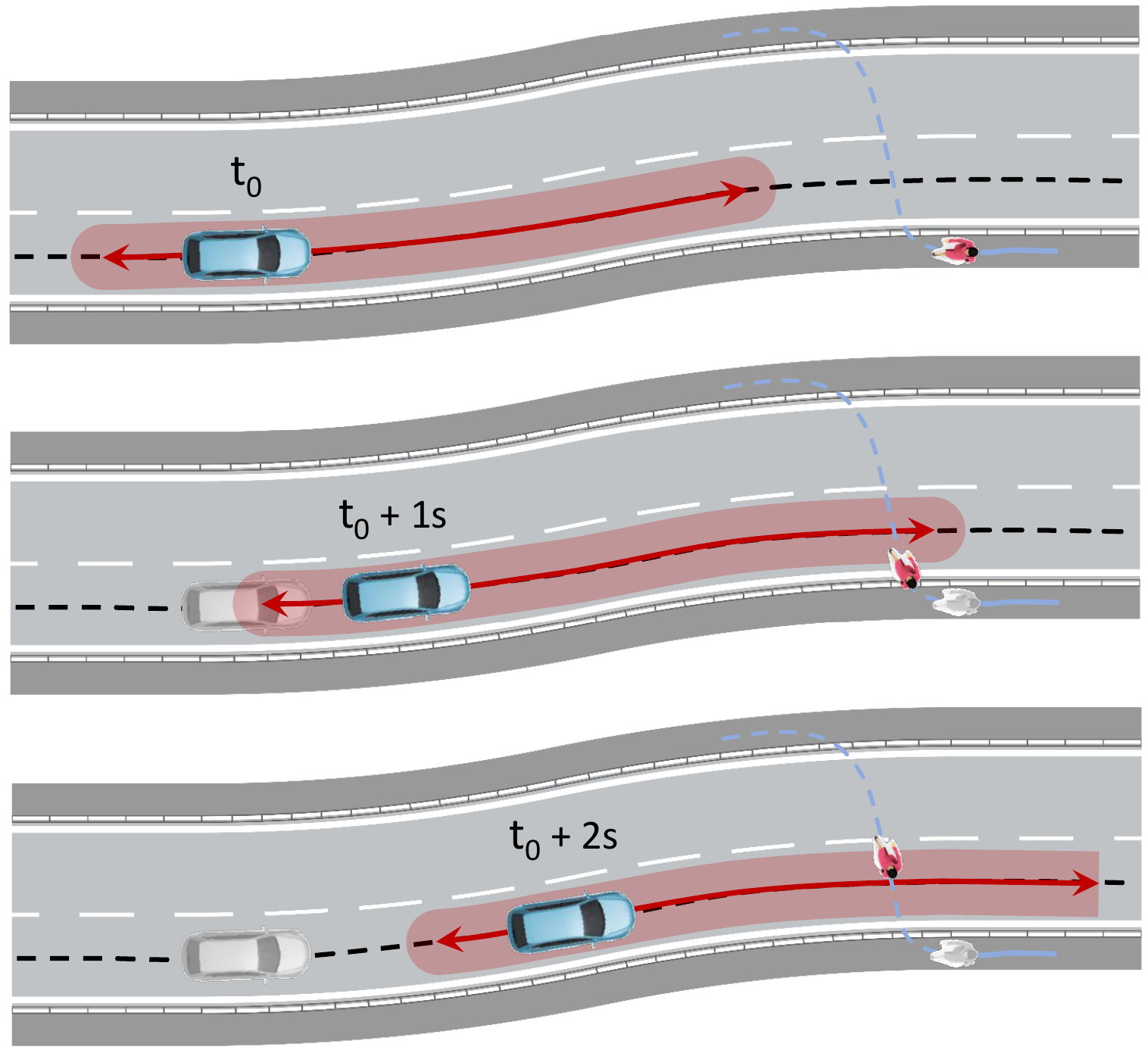}
		\label{fig:req_ADreactionPattern_woReaction}}	
	\hfil
	\subfloat[with system reaction]{\includegraphics[width=0.45\textwidth]{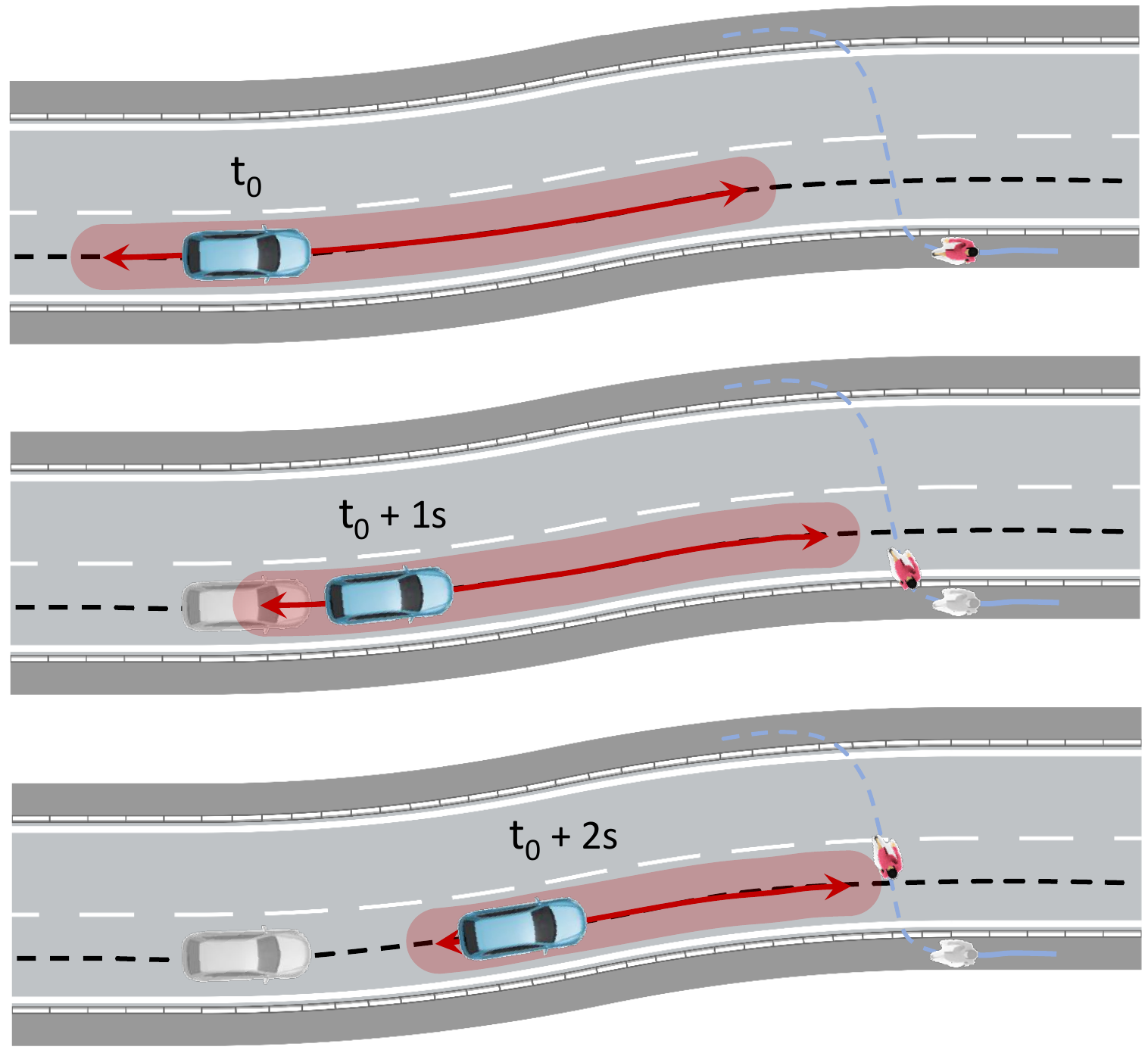}
		\label{fig:req_ADreactionPattern_wReaction}}
	\caption{Defining an appropriate system reaction based on the concept of a driver's comfort zone (red regions). In (a) the system first evaluates whether a pedestrian will violate the comfort zone in the future. In (b) a braking maneuver finally slows down the vehicle such that future comfort zone violations are circumvented. Dashed lines depict the paths of the vehicle and the pedestrian.}
	\label{fig:req_ADreactionPattern}
\end{figure*}

To realize an adequate system behavior we suggest that an automated vehicle monitors its future driving corridor, in particular with respect to violations of a defined minimum time gap. For this purpose, we define a comfort zone that extends along the vehicle's future path, as illustrated in Fig.~\ref{fig:req_ADreactionPattern}. The extent of the comfort zone thereby reflects a time gap which is considered comfortable by drivers and pedestrians. Consequently, its length depends on the vehicle's speed and can thus be adapted via braking (zone shrinks) or acceleration (zone is enlarged). In our implementation we choose a time gap of $3 \unit{s}$, which is slightly above the median value of the distribution in Fig.~\ref{fig:req_humanDriverAnalysis_TTC} and therefore corresponds to an average driving style. 

Furthermore, we propose that an automated vehicle evaluates whether its comfort zone is violated by a pedestrian -- currently or in the future. This is done by predicting the probability distribution of future pedestrian locations and intersecting them with the vehicle's future comfort zones (see Fig.~\ref{fig:req_ADreactionPattern_woReaction}). The latter are derived by shifting the current comfort zone along the planned path assuming constant vehicle velocity (i.e. no change in system behavior). In case of violations, the automated vehicle issues a system reaction, e.g. braking, such that the pedestrian's future trajectory does not violate the adapted comfort zones anymore (see Fig.~\ref{fig:req_ADreactionPattern_wReaction}).

From the above considerations it becomes evident that comfortable and anticipatory driving requires large prediction horizons. Pedestrians at least have to be predicted for a time horizon that corresponds to the comfort zone's time gap, i.e. $3 \unit{s}$ in our case. To realize natural driving behavior even larger horizons are required such that future comfort zone violations can be anticipated. However, with increasing prediction horizons the requirement on prediction accuracy can be gradually relaxed, since then an increasing time span is available to correct for an erroneous absence of a system reaction. On the other hand, to minimize false system reactions it is required that behavior planning takes prediction uncertainties into account. Pedestrian prediction models hence should yield uncertainty estimates (e.g. in terms of probability distributions over future pedestrian locations).

It should be noted that the comfort zone is not necessarily restricted to regions in front of the vehicle. Rather, it may also cover a region behind the vehicle. This allows to evaluate whether the vehicle can pass a pedestrian with a sufficient time gap before the pedestrian enters the driving corridor. Similarly, the comfort zone may be extended to include infrastructure elements (e.g. zebra crossings) such that country-specific traffic rules are taken into account.


\subsection{Prediction Performance Metric: In-ROI Sensitivity (IRS)}
\label{sec:roi_metric}

Building on the AD system reaction, we now
derive an application specific performance metric for our prediction model. To this end we define pedestrian behavior prediction as a binary classification task. 

\begin{figure}[ht]
	\centering
	\begin{minipage}[t]{1.0\linewidth}
		\centering
		\subfloat[]{\includegraphics[trim=0 0 0 0,clip,width=0.95\textwidth]{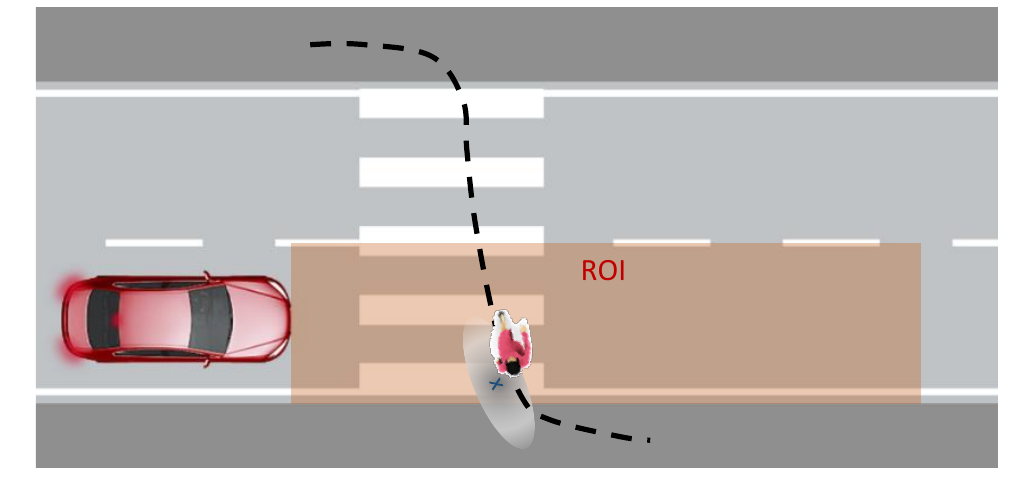}
			\label{fig:req_roiMetric_scene}}
		\vfil
		\subfloat[]{\includegraphics[trim=0 0 0 0,clip,width=\textwidth]{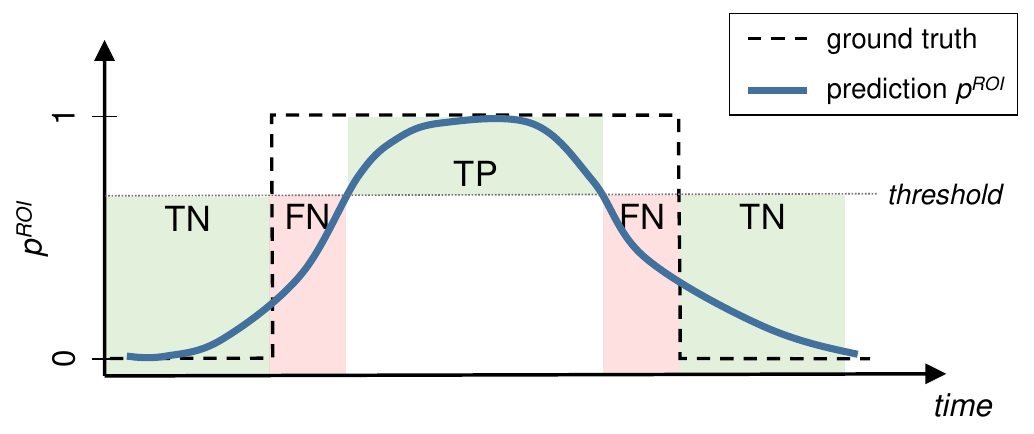}
			\label{fig:req_roiMetric_proi}}
	\end{minipage}
	\caption{Derivation of the proposed performance metric.
          (a) Ground truth pedestrian trajectory crossing the region of interest.
          (b) Prediction of in-ROI probability (solid) and ground truth of
          in-ROI state (dashed). For one recorded traffic scene, $3 \unit{s}$
          predictions of pedestrian and of ego vehicle ROI have been started from
          all possible points in time (x-axis).}
	\label{fig:req_roiMetric}
\end{figure}

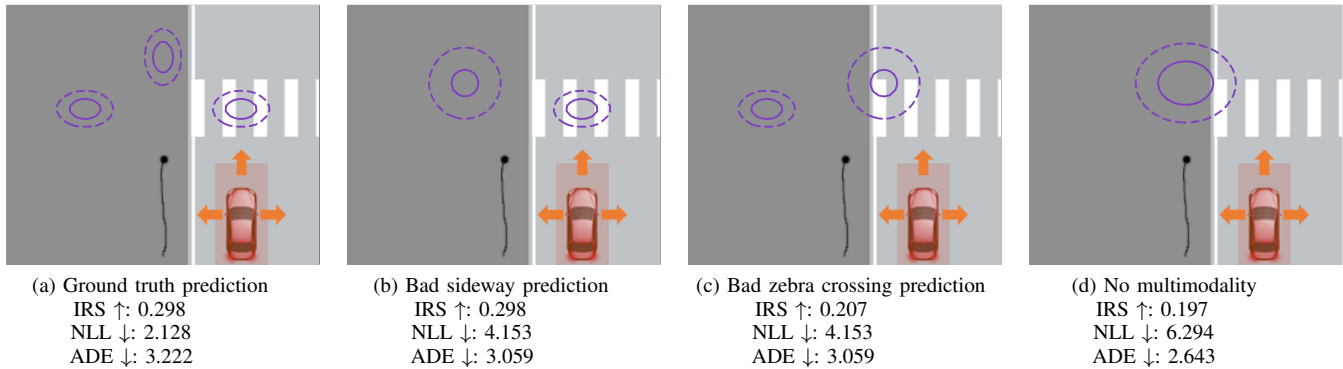
\begin{figure*}[ht!]
	\centering
	\subfloat[Ground truth prediction][\centering Ground truth prediction \newline IRS $\uparrow$: 0.298 \newline  NLL $\downarrow$: 2.128 \newline ADE $\downarrow$: 3.222 \newline ]{
\begin{tikzpicture}[scale=0.605]

\definecolor{color0}{rgb}{0.5,0.25,0.75}

\begin{axis}[
ticks=none,
x grid style={white!69.0196078431373!black},
xmin=-6, xmax=6,
y grid style={white!69.0196078431373!black},
ymin=-5, ymax=5,
]
\addplot graphics [includegraphics cmd=\pgfimage,xmin=-6, xmax=6, ymin=-5, ymax=5] {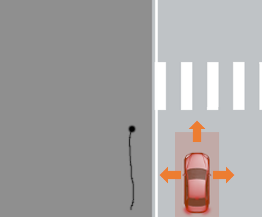};
\path [draw=color0, very thick]
(axis cs:-3.36734693877551,0.69546183623509)
--(axis cs:-3.16326530612245,0.626443762540176)
--(axis cs:-2.95918367346939,0.6125296447955)
--(axis cs:-2.75510204081633,0.646111430093802)
--(axis cs:-2.60397202010817,0.714285714285714)
--(axis cs:-2.55102040816327,0.752406303682663)
--(axis cs:-2.40780537090989,0.918367346938775)
--(axis cs:-2.425674008389,1.12244897959184)
--(axis cs:-2.55102040816327,1.24599118499204)
--(axis cs:-2.69398164587493,1.3265306122449)
--(axis cs:-2.75510204081633,1.35238866529136)
--(axis cs:-2.95918367346939,1.38807548166484)
--(axis cs:-3.16326530612245,1.37328917532221)
--(axis cs:-3.30325260994242,1.3265306122449)
--(axis cs:-3.36734693877551,1.29986007451087)
--(axis cs:-3.57142857142857,1.12343013765556)
--(axis cs:-3.57228126011322,1.12244897959184)
--(axis cs:-3.59193181624256,0.918367346938775)
--(axis cs:-3.57142857142857,0.890628709403746)
--(axis cs:-3.39968119880156,0.714285714285714)
--(axis cs:-3.36734693877551,0.69546183623509);

\path [draw=color0, very thick]
(axis cs:2.75510204081633,0.646111430093802)
--(axis cs:2.95918367346939,0.6125296447955)
--(axis cs:3.16326530612245,0.626443762540176)
--(axis cs:3.36734693877551,0.69546183623509)
--(axis cs:3.39968119880156,0.714285714285714)
--(axis cs:3.57142857142857,0.890628709403746)
--(axis cs:3.59193181624256,0.918367346938775)
--(axis cs:3.57228126011322,1.12244897959184)
--(axis cs:3.57142857142857,1.12343013765556)
--(axis cs:3.36734693877551,1.29986007451087)
--(axis cs:3.30325260994242,1.3265306122449)
--(axis cs:3.16326530612245,1.37328917532221)
--(axis cs:2.95918367346939,1.38807548166484)
--(axis cs:2.75510204081633,1.35238866529136)
--(axis cs:2.69398164587493,1.3265306122449)
--(axis cs:2.55102040816327,1.24599118499204)
--(axis cs:2.425674008389,1.12244897959184)
--(axis cs:2.40780537090989,0.918367346938775)
--(axis cs:2.55102040816327,0.752406303682663)
--(axis cs:2.60397202010817,0.714285714285714)
--(axis cs:2.75510204081633,0.646111430093802);

\path [draw=color0, very thick]
(axis cs:-0.102040816326531,2.41575422820441)
--(axis cs:0.102040816326531,2.41575422820441)
--(axis cs:0.245887035739489,2.55102040816327)
--(axis cs:0.306122448979592,2.64607385030264)
--(axis cs:0.353695975264047,2.75510204081633)
--(axis cs:0.388201463113119,2.95918367346939)
--(axis cs:0.373904622831779,3.16326530612245)
--(axis cs:0.306122448979592,3.36002045959852)
--(axis cs:0.302693182014369,3.36734693877551)
--(axis cs:0.116643220892896,3.57142857142857)
--(axis cs:0.10204081632653,3.58319027303939)
--(axis cs:-0.10204081632653,3.58319027303939)
--(axis cs:-0.116643220892896,3.57142857142857)
--(axis cs:-0.302693182014369,3.36734693877551)
--(axis cs:-0.306122448979592,3.36002045959852)
--(axis cs:-0.373904622831779,3.16326530612245)
--(axis cs:-0.388201463113119,2.95918367346939)
--(axis cs:-0.353695975264047,2.75510204081633)
--(axis cs:-0.306122448979592,2.64607385030264)
--(axis cs:-0.245887035739489,2.55102040816327)
--(axis cs:-0.102040816326531,2.41575422820441);

\path [draw=color0, very thick, dash pattern=on 5.55pt off 2.4pt]
(axis cs:-3.57142857142857,0.398047072521438)
--(axis cs:-3.36734693877551,0.341099688997489)
--(axis cs:-3.16326530612245,0.317350429699957)
--(axis cs:-2.95918367346939,0.312562553531819)
--(axis cs:-2.75510204081633,0.324118114039764)
--(axis cs:-2.55102040816327,0.358487284130814)
--(axis cs:-2.3469387755102,0.437639583141542)
--(axis cs:-2.23636137994464,0.510204081632653)
--(axis cs:-2.14285714285714,0.576263581330783)
--(axis cs:-2.00387129799555,0.714285714285714)
--(axis cs:-1.93877551020408,0.884344247004171)
--(axis cs:-1.92142820491111,0.918367346938775)
--(axis cs:-1.9348468069071,1.12244897959184)
--(axis cs:-1.93877551020408,1.12900253595093)
--(axis cs:-2.03539694398943,1.3265306122449)
--(axis cs:-2.14285714285714,1.42661483673595)
--(axis cs:-2.31344763999607,1.53061224489796)
--(axis cs:-2.3469387755102,1.55186620731901)
--(axis cs:-2.55102040816327,1.64613242063353)
--(axis cs:-2.75510204081633,1.68706428944517)
--(axis cs:-2.95918367346939,1.70082635252389)
--(axis cs:-3.16326530612245,1.69512424453928)
--(axis cs:-3.36734693877551,1.66684012934636)
--(axis cs:-3.57142857142857,1.59901879920095)
--(axis cs:-3.68870300493185,1.53061224489796)
--(axis cs:-3.77551020408163,1.48510643442193)
--(axis cs:-3.95625643119971,1.3265306122449)
--(axis cs:-3.97959183673469,1.28448898233128)
--(axis cs:-4.08114710402031,1.12244897959184)
--(axis cs:-4.09120045301524,0.918367346938775)
--(axis cs:-3.9836991445146,0.714285714285714)
--(axis cs:-3.97959183673469,0.709880299922785)
--(axis cs:-3.77551020408163,0.521222182995719)
--(axis cs:-3.75737345914153,0.510204081632653)
--(axis cs:-3.57142857142857,0.398047072521438);

\path [draw=color0, very thick, dash pattern=on 5.55pt off 2.4pt]
(axis cs:2.3469387755102,0.437639583141542)
--(axis cs:2.55102040816327,0.358487284130814)
--(axis cs:2.75510204081633,0.324118114039764)
--(axis cs:2.95918367346939,0.312562553531819)
--(axis cs:3.16326530612245,0.317350429699957)
--(axis cs:3.36734693877551,0.341099688997489)
--(axis cs:3.57142857142857,0.398047072521438)
--(axis cs:3.75737345914153,0.510204081632653)
--(axis cs:3.77551020408163,0.521222182995719)
--(axis cs:3.97959183673469,0.709880299922785)
--(axis cs:3.9836991445146,0.714285714285714)
--(axis cs:4.09120045301524,0.918367346938775)
--(axis cs:4.08114710402031,1.12244897959184)
--(axis cs:3.97959183673469,1.28448898233128)
--(axis cs:3.95625643119971,1.3265306122449)
--(axis cs:3.77551020408163,1.48510643442192)
--(axis cs:3.68870300493184,1.53061224489796)
--(axis cs:3.57142857142857,1.59901879920095)
--(axis cs:3.36734693877551,1.66684012934636)
--(axis cs:3.16326530612245,1.69512424453928)
--(axis cs:2.95918367346939,1.70082635252389)
--(axis cs:2.75510204081633,1.68706428944517)
--(axis cs:2.55102040816327,1.64613242063353)
--(axis cs:2.3469387755102,1.55186620731901)
--(axis cs:2.31344763999607,1.53061224489796)
--(axis cs:2.14285714285714,1.42661483673595)
--(axis cs:2.03539694398943,1.3265306122449)
--(axis cs:1.93877551020408,1.12900253595093)
--(axis cs:1.9348468069071,1.12244897959184)
--(axis cs:1.92142820491111,0.918367346938775)
--(axis cs:1.93877551020408,0.884344247004171)
--(axis cs:2.00387129799555,0.714285714285714)
--(axis cs:2.14285714285714,0.576263581330784)
--(axis cs:2.23636137994464,0.510204081632653)
--(axis cs:2.3469387755102,0.437639583141542);

\path [draw=color0, very thick, dash pattern=on 5.55pt off 2.4pt]
(axis cs:-0.102040816326531,1.92739746265859)
--(axis cs:0.102040816326531,1.92739746265859)
--(axis cs:0.122519467200721,1.93877551020408)
--(axis cs:0.306122448979592,2.01861734820778)
--(axis cs:0.425465072617763,2.14285714285714)
--(axis cs:0.510204081632653,2.27213102193552)
--(axis cs:0.558456401279968,2.3469387755102)
--(axis cs:0.64461389465558,2.55102040816327)
--(axis cs:0.682024831009207,2.75510204081633)
--(axis cs:0.694603090521986,2.95918367346939)
--(axis cs:0.68939147409313,3.16326530612245)
--(axis cs:0.663540339774157,3.36734693877551)
--(axis cs:0.60155295557205,3.57142857142857)
--(axis cs:0.510204081632653,3.72550891909349)
--(axis cs:0.482020433298632,3.77551020408163)
--(axis cs:0.306122448979592,3.96990125264982)
--(axis cs:0.286483965180868,3.97959183673469)
--(axis cs:0.102040816326531,4.08672822413799)
--(axis cs:-0.102040816326531,4.08672822413799)
--(axis cs:-0.286483965180868,3.97959183673469)
--(axis cs:-0.306122448979592,3.96990125264982)
--(axis cs:-0.482020433298632,3.77551020408163)
--(axis cs:-0.510204081632653,3.72550891909349)
--(axis cs:-0.601552955572051,3.57142857142857)
--(axis cs:-0.663540339774158,3.36734693877551)
--(axis cs:-0.68939147409313,3.16326530612245)
--(axis cs:-0.694603090521986,2.95918367346939)
--(axis cs:-0.682024831009207,2.75510204081633)
--(axis cs:-0.64461389465558,2.55102040816327)
--(axis cs:-0.558456401279969,2.3469387755102)
--(axis cs:-0.510204081632653,2.27213102193552)
--(axis cs:-0.425465072617763,2.14285714285714)
--(axis cs:-0.306122448979592,2.01861734820778)
--(axis cs:-0.122519467200721,1.93877551020408)
--(axis cs:-0.102040816326531,1.92739746265859);

\end{axis}

\end{tikzpicture}
		\label{fig:exp_toy_groundtruth}}
	\hfil
	\subfloat[Bad sideway prediction][\centering Bad sideway prediction \newline IRS $\uparrow$: 0.298 \newline  NLL $\downarrow$: 4.153 \newline ADE $\downarrow$: 3.059 \newline ]{
\begin{tikzpicture}[scale=0.605]

\definecolor{color0}{rgb}{0.5,0.25,0.75}

\begin{axis}[
ticks=none,
x grid style={white!69.0196078431373!black},
xmin=-6, xmax=6,
y grid style={white!69.0196078431373!black},
ymin=-5, ymax=5,
]
\addplot graphics [includegraphics cmd=\pgfimage,xmin=-6, xmax=6, ymin=-5, ymax=5] {images/exp_toyexample/BackgroundImage.png};
\path [draw=color0, very thick]
(axis cs:2.75510204081633,0.646111429339242)
--(axis cs:2.95918367346939,0.612529644799632)
--(axis cs:3.16326530612245,0.626443762653676)
--(axis cs:3.36734693877551,0.695461836397699)
--(axis cs:3.39968119852147,0.714285714285714)
--(axis cs:3.57142857142857,0.890628709696658)
--(axis cs:3.59193181602601,0.918367346938775)
--(axis cs:3.57228125988821,1.12244897959184)
--(axis cs:3.57142857142857,1.12343013739667)
--(axis cs:3.36734693877551,1.29986007434846)
--(axis cs:3.30325260962675,1.3265306122449)
--(axis cs:3.16326530612245,1.37328917527569)
--(axis cs:2.95918367346939,1.38807548212956)
--(axis cs:2.75510204081633,1.35238866896461)
--(axis cs:2.69398162582641,1.3265306122449)
--(axis cs:2.55102040816327,1.24599120877176)
--(axis cs:2.42567393980693,1.12244897959184)
--(axis cs:2.40780532238551,0.918367346938775)
--(axis cs:2.55102040816327,0.752406293724849)
--(axis cs:2.60397201043326,0.714285714285714)
--(axis cs:2.75510204081633,0.646111429339242);

\path [draw=color0, very thick]
(axis cs:-1.53061224489796,1.48240311020282)
--(axis cs:-1.3265306122449,1.5099986027378)
--(axis cs:-1.28621192067226,1.53061224489796)
--(axis cs:-1.12244897959184,1.65110455603827)
--(axis cs:-1.05698273394761,1.73469387755102)
--(axis cs:-0.987200081245431,1.93877551020408)
--(axis cs:-1.00421223544918,2.14285714285714)
--(axis cs:-1.11245767850169,2.3469387755102)
--(axis cs:-1.12244897959184,2.35754692573019)
--(axis cs:-1.3265306122449,2.48605031031825)
--(axis cs:-1.53061224489796,2.51707344670847)
--(axis cs:-1.73469387755102,2.45872959441348)
--(axis cs:-1.88046187682951,2.3469387755102)
--(axis cs:-1.93877551020408,2.2652102956356)
--(axis cs:-1.99753494932815,2.14285714285714)
--(axis cs:-2.01358680316933,1.93877551020408)
--(axis cs:-1.94774323460006,1.73469387755102)
--(axis cs:-1.93877551020408,1.72193667617286)
--(axis cs:-1.73469387755102,1.53535583534832)
--(axis cs:-1.72090206647017,1.53061224489796)
--(axis cs:-1.53061224489796,1.48240311020282);

\path [draw=color0, very thick, dash pattern=on 5.55pt off 2.4pt]
(axis cs:2.3469387755102,0.437639534880393)
--(axis cs:2.55102040816327,0.35848727941626)
--(axis cs:2.75510204081633,0.324118113510637)
--(axis cs:2.95918367346939,0.312562553493039)
--(axis cs:3.16326530612245,0.317350429732424)
--(axis cs:3.36734693877551,0.341099689052255)
--(axis cs:3.57142857142857,0.398047072605102)
--(axis cs:3.75737345900206,0.510204081632653)
--(axis cs:3.77551020408163,0.521222183080514)
--(axis cs:3.97959183673469,0.709880300096542)
--(axis cs:3.9836991443526,0.714285714285714)
--(axis cs:4.09120045290388,0.918367346938775)
--(axis cs:4.08114710390422,1.12244897959184)
--(axis cs:3.97959183673469,1.28448898214614)
--(axis cs:3.95625643109702,1.3265306122449)
--(axis cs:3.77551020408163,1.48510643433248)
--(axis cs:3.68870300476616,1.53061224489796)
--(axis cs:3.57142857142857,1.59901879910855)
--(axis cs:3.36734693877551,1.66684012932147)
--(axis cs:3.16326530612245,1.69512424473441)
--(axis cs:2.95918367346939,1.70082635400935)
--(axis cs:2.75510204081633,1.68706429951923)
--(axis cs:2.55102040816327,1.6461324917094)
--(axis cs:2.3469387755102,1.55186672810913)
--(axis cs:2.31344635452842,1.53061224489796)
--(axis cs:2.14285714285714,1.42661686538284)
--(axis cs:2.03539171929572,1.3265306122449)
--(axis cs:1.93877551020408,1.12901487811818)
--(axis cs:1.93483908617036,1.12244897959184)
--(axis cs:1.9214223423163,0.918367346938775)
--(axis cs:1.93877551020408,0.884335614811824)
--(axis cs:2.00386965150513,0.714285714285714)
--(axis cs:2.14285714285714,0.576263220855323)
--(axis cs:2.23636111759007,0.510204081632653)
--(axis cs:2.3469387755102,0.437639534880393);

\path [draw=color0, very thick, dash pattern=on 5.55pt off 2.4pt]
(axis cs:-1.93877551020408,0.697823830624)
--(axis cs:-1.73469387755102,0.639336227616192)
--(axis cs:-1.53061224489796,0.618414287810371)
--(axis cs:-1.3265306122449,0.629539096451087)
--(axis cs:-1.12244897959184,0.67562004536756)
--(axis cs:-1.02839613887047,0.714285714285714)
--(axis cs:-0.918367346938775,0.752587580503366)
--(axis cs:-0.714285714285714,0.870670127851264)
--(axis cs:-0.654161913661473,0.918367346938775)
--(axis cs:-0.510204081632653,1.04625480250821)
--(axis cs:-0.440371382509582,1.12244897959184)
--(axis cs:-0.306122448979592,1.3248878281233)
--(axis cs:-0.305037334857951,1.3265306122449)
--(axis cs:-0.201907549975944,1.53061224489796)
--(axis cs:-0.147241680803378,1.73469387755102)
--(axis cs:-0.125455636402633,1.93877551020408)
--(axis cs:-0.13076680692195,2.14285714285714)
--(axis cs:-0.16456088052601,2.3469387755102)
--(axis cs:-0.236011854565767,2.55102040816327)
--(axis cs:-0.306122448979592,2.67410285766671)
--(axis cs:-0.350186514840527,2.75510204081633)
--(axis cs:-0.510204081632653,2.95001744254668)
--(axis cs:-0.51899586396446,2.95918367346939)
--(axis cs:-0.714285714285714,3.12804255709848)
--(axis cs:-0.773609159193646,3.16326530612245)
--(axis cs:-0.918367346938775,3.25065085237502)
--(axis cs:-1.12244897959184,3.32506416349079)
--(axis cs:-1.3265306122449,3.36166797554381)
--(axis cs:-1.45631077865764,3.36734693877551)
--(axis cs:-1.53061224489796,3.37208021639798)
--(axis cs:-1.57112024712648,3.36734693877551)
--(axis cs:-1.73469387755102,3.35388574978114)
--(axis cs:-1.93877551020408,3.30742687082424)
--(axis cs:-2.14285714285714,3.21806568080816)
--(axis cs:-2.22717161425319,3.16326530612245)
--(axis cs:-2.3469387755102,3.08476480663742)
--(axis cs:-2.4844077404177,2.95918367346939)
--(axis cs:-2.55102040816327,2.88395229359108)
--(axis cs:-2.65234751249902,2.75510204081633)
--(axis cs:-2.75510204081633,2.55292537528147)
--(axis cs:-2.75615005267839,2.55102040816327)
--(axis cs:-2.83632088453876,2.3469387755102)
--(axis cs:-2.87423917785622,2.14285714285714)
--(axis cs:-2.88019852205463,1.93877551020408)
--(axis cs:-2.85575371385751,1.73469387755102)
--(axis cs:-2.79441643482913,1.53061224489796)
--(axis cs:-2.75510204081633,1.45023546204104)
--(axis cs:-2.70103551478921,1.3265306122449)
--(axis cs:-2.55446238128011,1.12244897959184)
--(axis cs:-2.55102040816326,1.11852838822217)
--(axis cs:-2.3469387755102,0.920925661937008)
--(axis cs:-2.3433994440933,0.918367346938775)
--(axis cs:-2.14285714285714,0.781147775948502)
--(axis cs:-1.97471617644489,0.714285714285714)
--(axis cs:-1.93877551020408,0.697823830624);

\end{axis}

\end{tikzpicture}
		\label{fig:exp_toy_modeleft}}
	\hfil
	\subfloat[Bad zebra crossing prediction][\centering Bad zebra crossing prediction \newline IRS $\uparrow$: 0.207 \newline  NLL $\downarrow$: 4.153 \newline ADE $\downarrow$: 3.059 \newline ]{
\begin{tikzpicture}[scale=0.605]

\definecolor{color0}{rgb}{0.5,0.25,0.75}

\begin{axis}[
ticks=none,
x grid style={white!69.0196078431373!black},
xmin=-6, xmax=6,
y grid style={white!69.0196078431373!black},
ymin=-5, ymax=5,
]
\addplot graphics [includegraphics cmd=\pgfimage,xmin=-6, xmax=6, ymin=-5, ymax=5] {images/exp_toyexample/BackgroundImage.png};
\path [draw=color0, very thick]
(axis cs:-3.36734693877551,0.695461836397702)
--(axis cs:-3.16326530612245,0.626443762653678)
--(axis cs:-2.95918367346939,0.612529644799634)
--(axis cs:-2.75510204081633,0.646111429339244)
--(axis cs:-2.60397201043326,0.714285714285714)
--(axis cs:-2.55102040816327,0.752406293724853)
--(axis cs:-2.40780532238552,0.918367346938775)
--(axis cs:-2.42567393980693,1.12244897959184)
--(axis cs:-2.55102040816327,1.24599120877176)
--(axis cs:-2.69398162582642,1.3265306122449)
--(axis cs:-2.75510204081633,1.35238866896461)
--(axis cs:-2.95918367346939,1.38807548212956)
--(axis cs:-3.16326530612245,1.37328917527569)
--(axis cs:-3.30325260962674,1.3265306122449)
--(axis cs:-3.36734693877551,1.29986007434845)
--(axis cs:-3.57142857142857,1.12343013739667)
--(axis cs:-3.57228125988821,1.12244897959184)
--(axis cs:-3.591931816026,0.918367346938775)
--(axis cs:-3.57142857142857,0.890628709696662)
--(axis cs:-3.39968119852147,0.714285714285714)
--(axis cs:-3.36734693877551,0.695461836397702);

\path [draw=color0, very thick]
(axis cs:1.3265306122449,1.50999860273781)
--(axis cs:1.53061224489796,1.48240311020283)
--(axis cs:1.72090206647015,1.53061224489796)
--(axis cs:1.73469387755102,1.53535583534833)
--(axis cs:1.93877551020408,1.72193667617287)
--(axis cs:1.94774323460005,1.73469387755102)
--(axis cs:2.01358680316933,1.93877551020408)
--(axis cs:1.99753494932814,2.14285714285714)
--(axis cs:1.93877551020408,2.26521029563559)
--(axis cs:1.8804618768295,2.3469387755102)
--(axis cs:1.73469387755102,2.45872959441347)
--(axis cs:1.53061224489796,2.51707344670846)
--(axis cs:1.3265306122449,2.48605031031824)
--(axis cs:1.12244897959184,2.35754692573018)
--(axis cs:1.1124576785017,2.3469387755102)
--(axis cs:1.00421223544919,2.14285714285714)
--(axis cs:0.987200081245438,1.93877551020408)
--(axis cs:1.05698273394762,1.73469387755102)
--(axis cs:1.12244897959184,1.65110455603828)
--(axis cs:1.28621192067228,1.53061224489796)
--(axis cs:1.3265306122449,1.50999860273781);

\path [draw=color0, very thick, dash pattern=on 5.55pt off 2.4pt]
(axis cs:-3.57142857142857,0.398047072605103)
--(axis cs:-3.36734693877551,0.341099689052256)
--(axis cs:-3.16326530612245,0.317350429732425)
--(axis cs:-2.95918367346939,0.31256255349304)
--(axis cs:-2.75510204081633,0.324118113510638)
--(axis cs:-2.55102040816327,0.358487279416261)
--(axis cs:-2.3469387755102,0.437639534880395)
--(axis cs:-2.23636111759007,0.510204081632653)
--(axis cs:-2.14285714285714,0.576263220855324)
--(axis cs:-2.00386965150513,0.714285714285714)
--(axis cs:-1.93877551020408,0.884335614811829)
--(axis cs:-1.92142234231631,0.918367346938775)
--(axis cs:-1.93483908617036,1.12244897959184)
--(axis cs:-1.93877551020408,1.12901487811818)
--(axis cs:-2.03539171929572,1.3265306122449)
--(axis cs:-2.14285714285714,1.42661686538284)
--(axis cs:-2.31344635452842,1.53061224489796)
--(axis cs:-2.3469387755102,1.55186672810913)
--(axis cs:-2.55102040816327,1.6461324917094)
--(axis cs:-2.75510204081633,1.68706429951923)
--(axis cs:-2.95918367346939,1.70082635400935)
--(axis cs:-3.16326530612245,1.69512424473441)
--(axis cs:-3.36734693877551,1.66684012932147)
--(axis cs:-3.57142857142857,1.59901879910855)
--(axis cs:-3.68870300476616,1.53061224489796)
--(axis cs:-3.77551020408163,1.48510643433248)
--(axis cs:-3.95625643109702,1.3265306122449)
--(axis cs:-3.97959183673469,1.28448898214613)
--(axis cs:-4.08114710390422,1.12244897959184)
--(axis cs:-4.09120045290388,0.918367346938775)
--(axis cs:-3.9836991443526,0.714285714285714)
--(axis cs:-3.97959183673469,0.709880300096545)
--(axis cs:-3.77551020408163,0.521222183080516)
--(axis cs:-3.75737345900205,0.510204081632653)
--(axis cs:-3.57142857142857,0.398047072605103);

\path [draw=color0, very thick, dash pattern=on 5.55pt off 2.4pt]
(axis cs:1.12244897959184,0.675620045367563)
--(axis cs:1.3265306122449,0.62953909645109)
--(axis cs:1.53061224489796,0.618414287810373)
--(axis cs:1.73469387755102,0.639336227616195)
--(axis cs:1.93877551020408,0.697823830624003)
--(axis cs:1.97471617644488,0.714285714285714)
--(axis cs:2.14285714285714,0.781147775948505)
--(axis cs:2.34339944409329,0.918367346938775)
--(axis cs:2.3469387755102,0.920925661937011)
--(axis cs:2.55102040816327,1.11852838822217)
--(axis cs:2.55446238128011,1.12244897959184)
--(axis cs:2.70103551478921,1.3265306122449)
--(axis cs:2.75510204081633,1.45023546204105)
--(axis cs:2.79441643482913,1.53061224489796)
--(axis cs:2.8557537138575,1.73469387755102)
--(axis cs:2.88019852205463,1.93877551020408)
--(axis cs:2.87423917785622,2.14285714285714)
--(axis cs:2.83632088453876,2.3469387755102)
--(axis cs:2.75615005267839,2.55102040816327)
--(axis cs:2.75510204081633,2.55292537528147)
--(axis cs:2.65234751249902,2.75510204081633)
--(axis cs:2.55102040816327,2.88395229359107)
--(axis cs:2.48440774041769,2.95918367346939)
--(axis cs:2.3469387755102,3.08476480663741)
--(axis cs:2.22717161425318,3.16326530612245)
--(axis cs:2.14285714285714,3.21806568080816)
--(axis cs:1.93877551020408,3.30742687082424)
--(axis cs:1.73469387755102,3.35388574978113)
--(axis cs:1.57112024712646,3.36734693877551)
--(axis cs:1.53061224489796,3.37208021639797)
--(axis cs:1.45631077865769,3.36734693877551)
--(axis cs:1.3265306122449,3.36166797554381)
--(axis cs:1.12244897959184,3.32506416349079)
--(axis cs:0.918367346938775,3.25065085237502)
--(axis cs:0.773609159193652,3.16326530612245)
--(axis cs:0.714285714285714,3.12804255709847)
--(axis cs:0.518995863964463,2.95918367346939)
--(axis cs:0.510204081632653,2.95001744254668)
--(axis cs:0.35018651484053,2.75510204081633)
--(axis cs:0.306122448979592,2.67410285766671)
--(axis cs:0.23601185456577,2.55102040816327)
--(axis cs:0.164560880526013,2.3469387755102)
--(axis cs:0.130766806921952,2.14285714285714)
--(axis cs:0.125455636402635,1.93877551020408)
--(axis cs:0.14724168080338,1.73469387755102)
--(axis cs:0.201907549975947,1.53061224489796)
--(axis cs:0.305037334857955,1.3265306122449)
--(axis cs:0.306122448979592,1.32488782812331)
--(axis cs:0.440371382509585,1.12244897959184)
--(axis cs:0.510204081632653,1.04625480250821)
--(axis cs:0.654161913661478,0.918367346938775)
--(axis cs:0.714285714285714,0.870670127851267)
--(axis cs:0.918367346938775,0.752587580503369)
--(axis cs:1.02839613887047,0.714285714285714)
--(axis cs:1.12244897959184,0.675620045367563);

\end{axis}

\end{tikzpicture}
		\label{fig:exp_toy_moderight}}
	\hfil
	\subfloat[No multimodality][\centering No multimodality \newline IRS $\uparrow$: 0.197 \newline  NLL $\downarrow$: 6.294 \newline ADE $\downarrow$: 2.643 \newline ]{
\begin{tikzpicture}[scale=0.605]

\definecolor{color0}{rgb}{0.5,0.25,0.75}

\begin{axis}[
ticks=none,
x grid style={white!69.0196078431373!black},
xmin=-6, xmax=6,
y grid style={white!69.0196078431373!black},
ymin=-5, ymax=5,
]
\addplot graphics [includegraphics cmd=\pgfimage,xmin=-6, xmax=6, ymin=-5, ymax=5] {images/exp_toyexample/BackgroundImage.png};
\path [draw=color0, very thick]
(axis cs:-0.510204081632653,1.26375593513085)
--(axis cs:-0.306122448979592,1.19558631311602)
--(axis cs:-0.102040816326531,1.1640726678517)
--(axis cs:0.102040816326531,1.1640726678517)
--(axis cs:0.306122448979592,1.19558631311602)
--(axis cs:0.510204081632653,1.26375593513085)
--(axis cs:0.627996548850247,1.3265306122449)
--(axis cs:0.714285714285714,1.38206032647514)
--(axis cs:0.87867342374331,1.53061224489796)
--(axis cs:0.918367346938775,1.58742304443669)
--(axis cs:1.00783935371371,1.73469387755102)
--(axis cs:1.05947686078074,1.93877551020408)
--(axis cs:1.04688827036601,2.14285714285714)
--(axis cs:0.966789211156375,2.3469387755102)
--(axis cs:0.918367346938775,2.41321462772487)
--(axis cs:0.798652845285707,2.55102040816327)
--(axis cs:0.714285714285714,2.6188039804965)
--(axis cs:0.510204081632653,2.73512088886692)
--(axis cs:0.454585358330463,2.75510204081633)
--(axis cs:0.306122448979592,2.8048324513498)
--(axis cs:0.102040816326531,2.83737965755619)
--(axis cs:-0.102040816326531,2.83737965755619)
--(axis cs:-0.306122448979592,2.8048324513498)
--(axis cs:-0.454585358330463,2.75510204081633)
--(axis cs:-0.510204081632653,2.73512088886692)
--(axis cs:-0.714285714285714,2.6188039804965)
--(axis cs:-0.798652845285707,2.55102040816327)
--(axis cs:-0.918367346938775,2.41321462772487)
--(axis cs:-0.966789211156375,2.3469387755102)
--(axis cs:-1.04688827036601,2.14285714285714)
--(axis cs:-1.05947686078074,1.93877551020408)
--(axis cs:-1.00783935371371,1.73469387755102)
--(axis cs:-0.918367346938775,1.58742304443669)
--(axis cs:-0.87867342374331,1.53061224489796)
--(axis cs:-0.714285714285714,1.38206032647514)
--(axis cs:-0.627996548850247,1.3265306122449)
--(axis cs:-0.510204081632653,1.26375593513085);

\path [draw=color0, very thick, dash pattern=on 5.55pt off 2.4pt]
(axis cs:-0.306122448979592,0.495604335580309)
--(axis cs:-0.102040816326531,0.474677519327164)
--(axis cs:0.102040816326531,0.474677519327164)
--(axis cs:0.306122448979592,0.495604335580309)
--(axis cs:0.376667932810052,0.510204081632653)
--(axis cs:0.510204081632653,0.529435748615733)
--(axis cs:0.714285714285714,0.577962236953432)
--(axis cs:0.918367346938775,0.655657765462347)
--(axis cs:1.02899275571478,0.714285714285714)
--(axis cs:1.12244897959184,0.759021443819763)
--(axis cs:1.3265306122449,0.895893239847016)
--(axis cs:1.35472559845722,0.918367346938775)
--(axis cs:1.53061224489796,1.07569093669476)
--(axis cs:1.57584986573522,1.12244897959184)
--(axis cs:1.72613051509918,1.3265306122449)
--(axis cs:1.73469387755102,1.34337287985625)
--(axis cs:1.83881764202579,1.53061224489796)
--(axis cs:1.90027910707014,1.73469387755102)
--(axis cs:1.92477340724653,1.93877551020408)
--(axis cs:1.91880199751105,2.14285714285714)
--(axis cs:1.88080693323384,2.3469387755102)
--(axis cs:1.80047378416011,2.55102040816327)
--(axis cs:1.73469387755102,2.65618821158643)
--(axis cs:1.67621096691712,2.75510204081633)
--(axis cs:1.53061224489796,2.92310992763131)
--(axis cs:1.49624782281979,2.95918367346939)
--(axis cs:1.3265306122449,3.10663003245366)
--(axis cs:1.24279458310884,3.16326530612245)
--(axis cs:1.12244897959184,3.24331026745493)
--(axis cs:0.918367346938775,3.34092094506716)
--(axis cs:0.841198820330934,3.36734693877551)
--(axis cs:0.714285714285714,3.42024293111228)
--(axis cs:0.510204081632653,3.47801917148539)
--(axis cs:0.306122448979592,3.51181691060986)
--(axis cs:0.102040816326531,3.52744102484175)
--(axis cs:-0.102040816326531,3.52744102484175)
--(axis cs:-0.306122448979592,3.51181691060986)
--(axis cs:-0.510204081632653,3.47801917148539)
--(axis cs:-0.714285714285714,3.42024293111228)
--(axis cs:-0.841198820330934,3.36734693877551)
--(axis cs:-0.918367346938775,3.34092094506716)
--(axis cs:-1.12244897959184,3.24331026745493)
--(axis cs:-1.24279458310884,3.16326530612245)
--(axis cs:-1.3265306122449,3.10663003245366)
--(axis cs:-1.49624782281979,2.95918367346939)
--(axis cs:-1.53061224489796,2.92310992763131)
--(axis cs:-1.67621096691712,2.75510204081633)
--(axis cs:-1.73469387755102,2.65618821158643)
--(axis cs:-1.80047378416011,2.55102040816327)
--(axis cs:-1.88080693323384,2.3469387755102)
--(axis cs:-1.91880199751105,2.14285714285714)
--(axis cs:-1.92477340724653,1.93877551020408)
--(axis cs:-1.90027910707014,1.73469387755102)
--(axis cs:-1.83881764202579,1.53061224489796)
--(axis cs:-1.73469387755102,1.34337287985625)
--(axis cs:-1.72613051509918,1.3265306122449)
--(axis cs:-1.57584986573522,1.12244897959184)
--(axis cs:-1.53061224489796,1.07569093669476)
--(axis cs:-1.35472559845722,0.918367346938775)
--(axis cs:-1.3265306122449,0.895893239847016)
--(axis cs:-1.12244897959184,0.759021443819763)
--(axis cs:-1.02899275571478,0.714285714285714)
--(axis cs:-0.918367346938775,0.655657765462347)
--(axis cs:-0.714285714285714,0.577962236953432)
--(axis cs:-0.510204081632653,0.529435748615733)
--(axis cs:-0.376667932810052,0.510204081632653)
--(axis cs:-0.306122448979592,0.495604335580309);

\end{axis}

\end{tikzpicture}
		\label{fig:exp_toy_onemode}}
	\caption{Toy example for metric assessment: The toy example highlights differences between our
		metrics in a crossing scene. Background colors specify different ground types, where a pedestrian (black line) is walking on a sidewalk (dark gray) close to a street (bright gray) towards a zebra crossing (violet). (a) corresponds to a three-modal ground truth distribution, (b) represents a model with inaccurate on-sidewalk prediction, (c) shows a model with inaccurate crossing prediction, while (d) does not capture multiple modes at all. The captions contain corresponding metrics when comparing the predictions (a) - (d) to the ground truth distribution (a). For computing IRS, we have varied the positions of car ROIs on the street.}
	\label{fig:exp_metric_toy}
\end{figure*}

Fig.~\ref{fig:req_roiMetric_scene} shows a typical traffic scene with the
trajectory of a crossing pedestrian as well as the ego vehicle's comfort zone. The latter is referred to as Region of Interest (ROI) in the following. The task of the AD system is to anticipate predicted violations of the ROI by pedestrians, i.e. a classification whether a pedestrian will be located inside the ROI in the future. We define the in-ROI probability $P^{\ROI}_{t+T}$ at time $t+T$ as
\begin{equation}
P^{\ROI}_{t+T} = \int_{x_{t+T} \in \ROI_{t+T}} p(x_{t+T}|x_{0:t},C)~dx_{t+T},
\end{equation}
where $\ROI_{t+T}$ denotes the predicted ROI and $p(x_{t+T}|x_{t-H+1:t},C)$ the predictive distribution of pedestrian locations given the pedestrian's past trajectory $x_{t-H+1:t}$ and contextual cues $C$. Computing the in-ROI probability requires integration of the predictive distribution over the ROI. This can be approximated e.g. with a Monte-Carlo approach using samples from the distribution.

Fig.~\ref{fig:req_roiMetric_proi} illustrates the evolvement of the scene and the prediction over time. Specifically, the dashed line represents the true state of the pedestrian with regard to being inside the ROI for a given prediction horizon, e.g. $T = 3\unit{s}$, whereas the solid blue line shows the predicted in-ROI probability for this prediction horizon. Thresholding $P^{\ROI}_{t+T}$ finally yields an in-ROI classification that is compared to the true state for each sample $t$.

Thus, we have defined a classification problem (with predicted class probabilities) for which textbook metrics can be applied. In particular, we choose the True Positive Rate (TPR) and the False Positive Rate (FPR) which are defined as
\begin{eqnarray}
\TPR & = & \TP / (\TP + \FN) \\
\FPR & = & \FP / (\FP + \TN),
\end{eqnarray}
where $\TP$, $\TN$, $\FP$, and $\FN$ denote the number of true positive, true negative, false positive, and false negative samples, respectively. For metric calculation, we accumulate classification results from the set of test samples that are relevant for the AD system reaction. In our evaluation, we consider samples with $\TTC < 5\unit{s}$ as relevant, which excludes samples that will not result in a system reaction as defined in section \ref{sec:ad_system_reaction}.

In general, there is a trade off between TPR and FPR that can be represented as a ROC curve, where points on the curve are generated by varying the classification threshold. Fig.~\ref{fig:res_roc} shows ROC curves for different prediction horizons.
The models and the confidence bands shown in the figure will be discussed in Sec.~\ref{sec:results}.
These curves  allow for studying said trade off which would be difficult to assess from requirements given a priori. In particular, we use ROC curves to determine a FPR for each prediction horizon according to the following reasoning.

For an application of prediction models in an AD system, the performance (TPR) at low FPR is of particular interest, since high FPR would result in an unacceptable number of false system reactions. However, acceptable FPRs differ with respect to prediction horizons. Small prediction horizons potentially correspond to critical situations that require a rather strong reaction by the automated vehicle. In such situations an erroneous system reaction is highly unacceptable and potentially dangerous. On the other hand, large prediction horizons correspond to situations that are still uncritical but which are relevant for anticipatory driving. These situations require much weaker reactions by the automated vehicle and therefore erroneous system reactions are more acceptable. Hence, we allow for larger FPRs for long-term prediction compared to very small FPRs in the short-term prediction case. Specifically, we consider FPRs of 2.5\%, 5\%, 10\%, and 15\% as suitable working 
points for the 1\unit{s}, 2\unit{s}, 3\unit{s}, and 4\unit{s} predictions, 
respectively. Thus, we propose In-ROI Sensitivity (IRS) as a prediction performance metric, which measures the In-ROI TPR at the respective FPRs for the different prediction horizons.

\subsection{Metric Assessment}
\label{sec:exp_metric_assessment}
In the previous section, we derive an application-specific IRS-metric with an intuitive function-level interpretation. We postulate that it is preferable to traditional metrics for pedestrian future prediction, as it focuses on aspects of the prediction, which affect the corresponding AD system reaction. To highlight differences between the proposed metric and traditional ones, we perform a metric assessment based on toy data. The toy example is illustrated in Fig.~\ref{fig:exp_metric_toy}, where background colors specify different ground types. It corresponds to a crossing scene, where a pedestrian (black line) is walking on a sidewalk (dark gray) close to a street (bright gray) towards a zebra crossing (white). We visualize distributions with violet contour plots, where Fig.~\ref{fig:exp_toy_groundtruth} corresponds to the three-modal ground truth distribution, Fig.~\ref{fig:exp_toy_modeleft} illustrates a model with inaccurate on-sidewalk prediction, Fig.~\ref{fig:exp_toy_moderight} shows a model with inaccurate crossing prediction, while Fig.~\ref{fig:exp_toy_onemode} does not capture multiple modes at all. The captions contain corresponding metrics when comparing the prediction to the ground truth distribution in Fig.~\ref{fig:exp_toy_groundtruth}. For computing IRS, we have varied the positions of ego vehicle ROIs on the street.

We observe that matching the ground truth distribution yields highest NLL, while wrong predictions always lead to worse scores. Interestingly, ADE results in contradictory findings, as it favors uni-modal distributions due to measuring expected Euclidean error. Finally, according to our IRS metric, two predictions lead to similarly good results: Fig.~\ref{fig:exp_toy_groundtruth}, which matches the ground truth distribution correctly, and Fig.~\ref{fig:exp_toy_modeleft}, which only matches the relevant road / zebra mode correctly. Consequently, the IRS metric focuses on relevant aspects of the prediction, while not evaluating irrelevant parts (e.g. accurate prediction of pedestrians on sidewalks). This helps in finding the right balance between model complexity and predictive performance.


\section{Pedestrian Prediction Model}
\label{sec:rnnModel}
In this section, we outline a simple yet powerful enough pedestrian prediction model for assessing our proposed metric and the relevance of different feature combinations. Section~\ref{sec:requirements} analyzes human driving behavior and derives an expected AD system reaction pattern. Such a reaction pattern requires judging whether the probability of a pedestrian being inside a future comfort zone of the ego vehicle exceeds a certain threshold. Based on this reaction pattern, we can derive desired properties of prediction models:
\begin{itemize}
	\item \emph{Prediction of a continuous distribution over future pedestrian locations} for computing the likelihood of a pedestrian being inside the future comfort zone of the ego vehicle. 
	\item \emph{Predictive distributions over different prediction horizons} to evaluate comfort zone violations for different prediction horizons.
	\item \emph{Ability to model complex multi-modal distributions} as future behavior can have several non-trivial modes (e.g. going straight or crossing the street).
	\item \emph{Learning influences of contextual cues} that affect pedestrian behavior. Concretely, we focus on pedestrian motion and poses, the static map, and interactions with the ego vehicle, as we expect that these features have the strongest influence on the AD system reaction.
\end{itemize}
Due to the limited prediction horizon and only a subset of pedestrians being relevant for the IRS metric, we believe that single agent prediction models that condition on static and dynamic contextual cues are sufficient for our purpose. 


\subsection{Model and Feature Integration}
In Section~\ref{sec:relatedWork}, we refer to several recent approaches for pedestrian behavior prediction, which differ in terms of model types, architectures, learning method, or availability of features. However, most recent approaches use deep learning based models for learning complex, non-linear functional dependencies from input features to the predicted behavior. Especially Conditional Variational Autoencoders (CVAEs) \cite{sohn2015learning} and Conditional Generative Adversarial Networks (CGANs) \cite{mirza2014conditional} have proven to be suitable for such use cases as they can learn complex, continuous distributions by introducing continuous latent variables. Furthermore, they allow to incorporate feature observations by conditioning on them. We specifically focus on CVAEs due to two reasons: their explicit density modeling, which allows to evaluate the NLL, as well as not suffering from mode collapse, which ensures to capture non-trivial modes in real human behavior.

\begin{figure}[t!]
	\centering
	\includegraphics[width=1.0\linewidth]{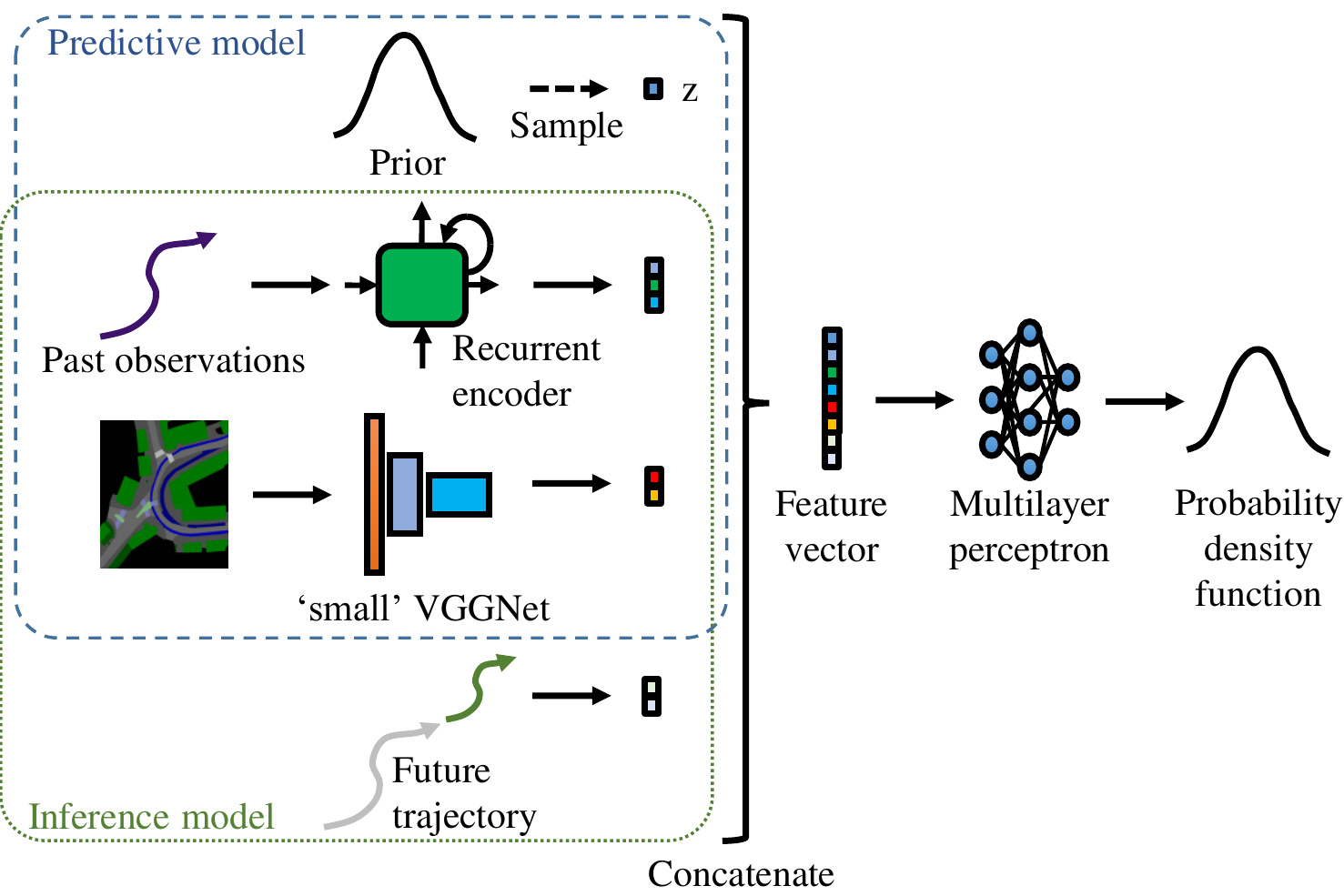}
	\caption{DL-based architecture of the pedestrian prediction CVAE. For inferring the latent variable, the inference model uses encoded past observations of a pedestrian, encoded static context of the environment, as well as the observed future trajectory. In contrast, the predictive model does not have access to the future trajectory, but predicts future locations given past observation, static context, and a sample of the latent variable $z$. The predictive and the inference model share the same encoders of past observations and static context. 
	}
	\label{fig:CVAE_architecture}
\end{figure}

In the following, we give an overview on the proposed CVAE architecture, with a focus on simplicity and flexibility for enabling feature relevance assessment with different sets of features under the proposed metric.
As indicated in Fig.~\ref{fig:CVAE_architecture} the model conditions on both static map information at a particular time step $s_t$ as well as dynamic features of the pedestrian $x_{t-H+1:t}$ over the last $H$ time steps. We encode dynamic features of the agent $x_{t-H+1:t}$ (e.g. relative motion between two time steps, head pose, body pose, distance to the ego vehicle) via recurrent encoders into an embedding space. In addition, we represent the static environment around the pedestrian via a grid in bird's-eye view, where different colors indicate different semantics (e.g. sidewalk, road, zebra crossing, building, isle, bicycle lane, unknown). These bird's-eye view grids are encoded via a small VGGNet architecture \cite{Simonyan2015VGG} to provide a static environment embedding vector. 

A CVAE comprises two models: The predictive model $p_\theta(x_{t+1:T}|x_{t-H+1:t},s_t,z)$ (dashed blue rectangle) predicts an 8-dimensional Gaussian distribution over the future pedestrian locations in $x$ and $y$ at four prediction time steps (1s, 2s, 3s, 4s) conditioned on past observations and a latent variable sample $z$ from the $Z$-dimensional multivariate standard Gaussian prior $p(z)$. The inference model $q_\phi(z|x_{t+1:T},x_{t-H+1:t},s_t)$ (dotted green rectangle) infers the latent variable $z$ from all available observations (future and past) and is only used during training. Additional implementation details can be found in the Appendix in Sec.~\ref{sec:implementationDetails}.

We train the model in the usual way by minimizing the evidence lower bound (ELBO): 
\begin{equation}
\E_{q_\phi(z|x_{t-H:T},s_t)}\left[ \log p_\theta(x_{t+1:T}|x_{t-H:t},s_t,z) \right] - D_{KL}(q_\phi || p(z))
\label{eq:CVAE_ELBO}
\end{equation}
The ELBO comprises a reconstruction term that aims to maximize the expected likelihood of future observations under the latent posterior distribution, as well as a Kullback-Leibler (KL) divergence term that tries to enhance agreement between prior and posterior latent distribution. So while the reconstruction term serves the purpose of obtaining an accurate prediction of the future trajectory given the observed past, the KL term acts as a regularizer. 

\subsection{Feature combinations:}
The proposed architecture allows for flexible changes of input features and enables an ablation study regarding the influence of different features on the prediction performance. Table~\ref{tab:features} denotes the features that we evaluate in this study. The most basic model only conditions on the past motion of the pedestrian and does not use any additional information. Hence, it can only learn future predictions based on cues in the pedestrian trajectory itself.
\begin{table}[ht!]
	\centering
	\caption{Description of features used in the CVAE.}
	\setlength{\tabcolsep}{5pt}
	\begin{tabularx}{\linewidth}{lX@{\hspace{10pt}}}
		\textbf{Feature} & \textbf{Description} \\
		Motion         	& Trajectory encoded as relative motion ($\Delta x, \Delta y$) between time steps.   		\\ 
		Egodist 		& Distance (x, y) to ego vehicle at every time step. \\
		Head         	& Head pose of the pedestrian at every time step.           								\\ 
		Body         	& Body pose of the pedestrian at every time step.      										\\ 
		Map         	& Semantic map around the last position of the pedestrian. 	\\ 
	\end{tabularx}
	\label{tab:features}
\end{table}


\section{Experimental Results}
\label{sec:results}


In the following, we evaluate the importance of different sets of contextual cues in terms of our proposed IRS metric. Furthermore, we highlight differences in the respective conclusions when comparing to traditional metrics.

\subsection{Experimental Setup}
Training, validation, and test subsets are drawn stratified from the three round courses in our dataset. The dataset is split by first assembling the test set. In order to be able to evaluate how pedestrian features like head pose affect the prediction performance, the whole test set needs to consist of labeled tracks. With the fraction of labeled tracks in the dataset being comparatively small, the test set has to be limited in size to ensure enough labeled tracks are available for training. We decided to constrain the test set size to 500 tracks while at the same time guaranteeing a large variety of scenes. This is achieved by drawing from the labeled subset via stratified random sampling, with each category, e.g. crossing/not crossing, distance from ego vehicle, crossing in front/behind the ego vehicle, being represented according to its proportion in the complete data set.

For training and validation, a minimum track length of 5s is required to enable training of prediction horizons of up to 4s. 5\% of the tracks fulfilling this criterion are randomly assigned to the validation set, the remaining tracks form the training set. Training and validation set contain 42,551 (7,175 labeled) and 2,278 (389 labeled), respectively.

In order to assess the variance of the test results, evaluations are
repeated using the bootstrap method \cite{Efron-Tibshirani-1993} with
$B=10000$ replications. Each replication uses an artificial test set
created from the original test set by resampling 500 tracks with
replacement. The bootstrap method produces an estimate of the
variability in the results that would be seen with completely new test
data taken from the same ground truth distribution. From the bootstrap replications we compute 50\% confidence intervals for the metrics IRS, NLL, and ADE, using the ``BCa'' method \cite{Hall-1988}, which corrects for bias and skewness of the sampling distribution.


\subsection{Quantitative Evaluation}

In order to introduce the procedure and to gain some intuition we first show evaluation results for two models in comparison.
One is a CVAE model that solely uses a pedestrian's past trajectory 
(\textit{CVAE motion}), while the other one is a CVAE model that employs a full feature 
set (\textit{CVAE full}) consisting of the pedestrian's past trajectory, 
their head and body pose, the distance to the ego vehicle's 
position, and the semantic map. In this way the effect of contextual cues on the prediction 
performance can be assessed. We run both models on the test dataset 
and evaluate the predictions with the system-specific ROI-metric
proposed in Sec.~\ref{sec:roi_metric}, using a range of FPR values.

\begin{figure}[ht]
  \centering
  \includegraphics[width=\linewidth]{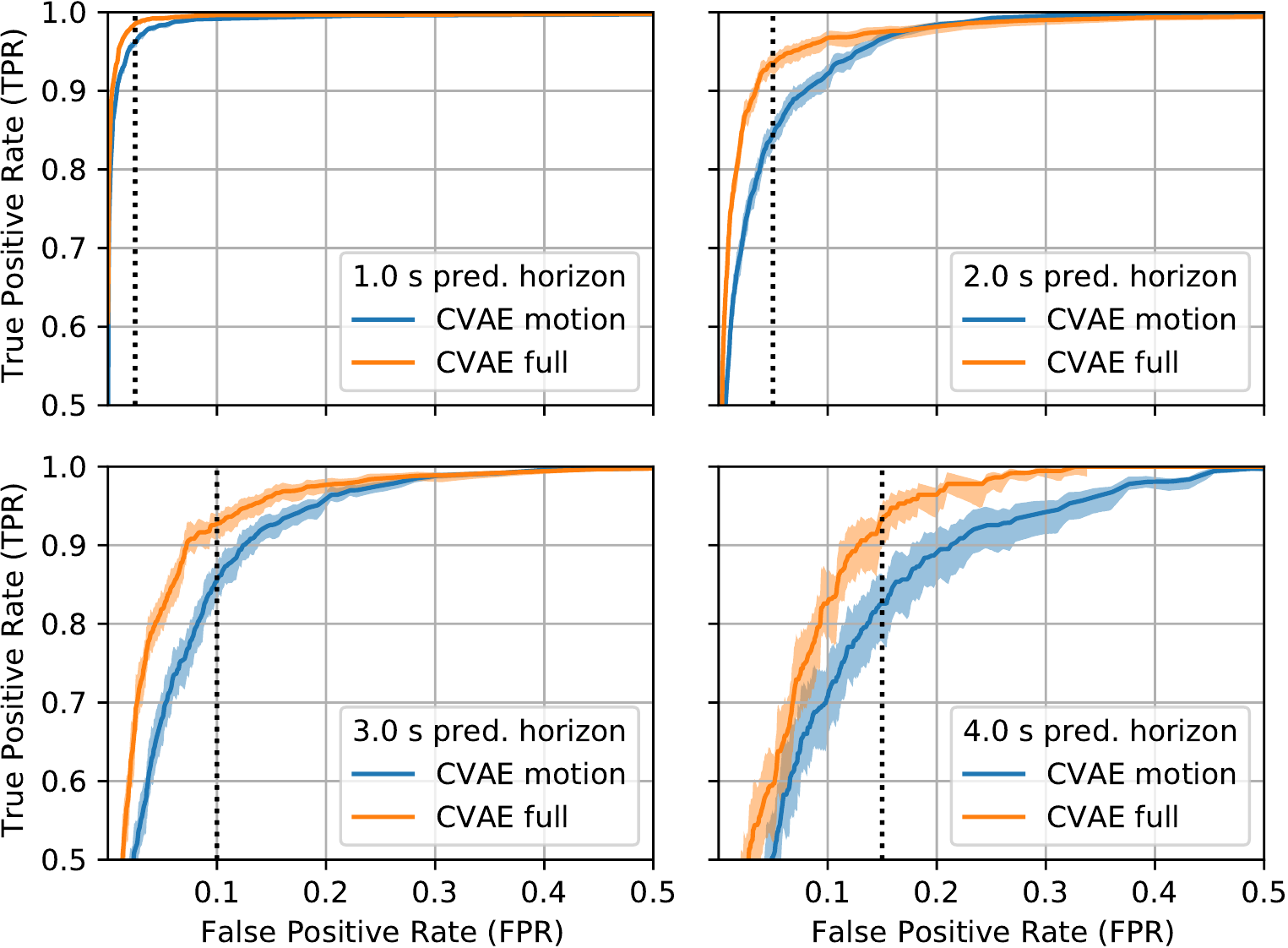}
       \caption{ROI-based metric results of the baseline CVAE motion model (pedestrian motion feature only)
                and of the full CVAE model (all features) for different prediction horizons.
		The shaded bands denote the TPR confidence interval at each FPR. Dashed lines represent the target FPR values chosen to define the IRS metric, see Sec.~\ref{sec:roi_metric}.}
	\label{fig:res_roc}
\end{figure}

The ROC curves for different prediction horizons are
depicted in Fig.~\ref{fig:res_roc} (solid line: TPR per FPR,
shaded area: 50\% confidence interval per FPR).
It becomes evident that the TPRs 
of the models decrease with increasing prediction horizons. For a prediction
horizon of 1\unit{s} both models achieve very good TPRs with the full CVAE 
model slightly outperforming the simpler one. This result confirms 
that for short-term pedestrian prediction the simpler model is already
well suited in terms of our IRS metric, and that contextual cues only have a minor benefit. In contrast,
for prediction horizons of 2 to 4\unit{s}, the full CVAE model strongly
outperforms the CVAE motion model. Our results thus confirm the importance of using contextual cues 
for long-term pedestrian prediction. A more detailed analysis of feature relevance follows in Sec.~\ref{sec:exp_feature_relevance}.

With longer prediction horizons the confidence bands become wider. This results from the fact that there is less data available for longer prediction horizons in our data set. We visualize the chosen FPRs from Sec. \ref{sec:requirements} as vertical dotted lines in Figure~\ref{fig:res_roc}. Our prediction performance metric IRS is equal to the TPR at the chosen FPRs. The IRS as well as the corresponding 50\% confidence intervals are reported in Table~\ref{tab:res_featureAssessment} for eight CVAE models with different input feature combinations.


\subsection{Feature Relevance Assessment}
\label{sec:exp_feature_relevance}

\begin{table*}[ht!]
	\centering
	\caption{Ablation Study of contextual cues}
	\setlength{\tabcolsep}{5pt}
	\begin{tabular}{clc@{\hspace{10pt}}ccccc@{\hspace{10pt}}ccccc@{\hspace{10pt}}cccc}
		\toprule
		& CVAE Model & & \multicolumn{4}{c}{IRS (\unit{\%}) $\uparrow$} & & \multicolumn{4}{c}{NLL $\downarrow$} & & \multicolumn{4}{c}{ADE (\unit{m}) $\downarrow$ }\\
		& & & 1\unit{s} & 2\unit{s} & 3\unit{s} & 4\unit{s} & & 1\unit{s} & 2\unit{s} & 3\unit{s} & 4\unit{s} & & 1\unit{s} & 2\unit{s} & 3\unit{s} & 4\unit{s} \\
		\toprule
		
		1 & motion &   &           \metric{96.1}{95.5}{96.5} &           \metric{84.6}{82.8}{86.1} &           \metric{85.4}{82.6}{87.8} &           \metric{82.6}{77.4}{86.3} &   &              \metric{0.13}{0.08}{0.18} &           \metric{1.54}{1.46}{1.59} &           \metric{2.39}{2.28}{2.47} &           \metric{3.10}{2.92}{3.21} &   &           \metric{0.53}{0.52}{0.54} &           \metric{1.10}{1.08}{1.12} &           \metric{1.73}{1.70}{1.77} &           \metric{2.48}{2.41}{2.54} \\
		\midrule
		2 & motion+egodist &   &           \metric{97.4}{96.9}{97.8} &           \metric{88.0}{86.4}{89.6} &           \metric{89.0}{86.9}{91.1} &           \metric{82.2}{76.8}{85.3} &   &           \metric{-0.14}{-0.20}{-0.10} &           \metric{1.27}{1.19}{1.33} &           \metric{2.14}{2.02}{2.22} &           \metric{2.88}{2.69}{2.99} &   &           \metric{0.49}{0.48}{0.50} &           \metric{1.03}{1.01}{1.05} &           \metric{1.64}{1.61}{1.68} &           \metric{2.39}{2.32}{2.45} \\
		\midrule
		3 & motion+head\&body &   &           \metric{97.7}{97.3}{98.2} &           \metric{89.7}{88.2}{91.3} &           \metric{89.1}{87.1}{91.0} &           \metric{84.6}{80.1}{88.2} &   &              \metric{0.05}{0.00}{0.09} &           \metric{1.45}{1.39}{1.50} &           \metric{2.32}{2.22}{2.39} &           \metric{3.04}{2.88}{3.13} &   &           \metric{0.51}{0.50}{0.52} &           \metric{1.09}{1.07}{1.11} &           \metric{1.74}{1.70}{1.77} &           \metric{2.49}{2.43}{2.55} \\
		\midrule
		4 & motion+egodist+head\&body &   &           \metric{98.1}{97.7}{98.4} &           \metric{91.0}{89.4}{92.2} &           \metric{90.8}{88.5}{92.1} &           \metric{86.8}{82.2}{89.7} &   &           \metric{-0.18}{-0.23}{-0.13} &           \metric{1.23}{1.15}{1.29} &           \metric{2.09}{1.99}{2.17} &           \metric{2.81}{2.62}{2.91} &   &           \metric{0.47}{0.46}{0.48} &           \metric{0.99}{0.97}{1.02} &           \metric{1.58}{1.55}{1.62} &           \metric{2.29}{2.23}{2.35} \\
		\midrule
		5 & motion+map &   &           \metric{97.9}{97.4}{98.3} &           \metric{90.7}{89.1}{92.1} &           \metric{90.6}{88.8}{92.6} &           \metric{91.2}{88.1}{93.7} &   &           \metric{-0.06}{-0.11}{-0.02} &           \metric{1.25}{1.18}{1.30} &           \metric{2.04}{1.94}{2.11} &           \metric{2.71}{2.57}{2.81} &   &           \metric{0.50}{0.49}{0.52} &           \metric{1.01}{0.99}{1.04} &           \metric{1.55}{1.52}{1.59} &           \metric{2.19}{2.13}{2.25} \\
		\midrule
		6 & motion+map+egodist &   &           \metric{98.0}{97.6}{98.4} &           \metric{92.2}{90.5}{93.3} &           \metric{90.4}{88.5}{92.5} &           \metric{87.2}{82.3}{89.8} &   &           \metric{-0.24}{-0.29}{-0.19} &           \metric{1.10}{1.02}{1.16} &           \metric{1.91}{1.79}{2.00} &           \metric{2.61}{2.42}{2.72} &   &           \metric{0.47}{0.46}{0.48} &           \metric{0.96}{0.94}{0.98} &           \metric{1.50}{1.46}{1.53} &           \metric{2.13}{2.06}{2.19} \\
		\midrule
		7 & motion+map+head\&body &   &           \metric{98.0}{97.6}{98.3} &           \metric{92.2}{90.6}{93.4} &           \metric{92.6}{91.1}{94.1} &  \metric{\mathbf{93.4}}{90.7}{95.8} &   &           \metric{-0.11}{-0.15}{-0.07} &           \metric{1.19}{1.13}{1.25} &           \metric{1.98}{1.89}{2.05} &           \metric{2.69}{2.56}{2.77} &   &           \metric{0.48}{0.47}{0.49} &           \metric{0.97}{0.95}{1.00} &           \metric{1.50}{1.47}{1.54} &           \metric{2.16}{2.10}{2.22} \\
		\midrule
		8 & motion+map+egodist+head\&body &   &  \metric{\mathbf{98.5}}{98.2}{98.9} &  \metric{\mathbf{93.5}}{92.1}{94.5} &  \metric{\mathbf{92.7}}{90.8}{94.0} &           \metric{93.3}{90.0}{95.3} &   &  \metric{\mathbf{-0.26}}{-0.31}{-0.21} &  \metric{\mathbf{1.05}}{0.98}{1.11} &  \metric{\mathbf{1.86}}{1.76}{1.94} &  \metric{\mathbf{2.55}}{2.37}{2.65} &   &  \metric{\mathbf{0.47}}{0.46}{0.48} &  \metric{\mathbf{0.95}}{0.93}{0.97} &  \metric{\mathbf{1.47}}{1.44}{1.51} &  \metric{\mathbf{2.12}}{2.05}{2.18} \\
		
		\bottomrule
	\end{tabular}
	\label{tab:res_featureAssessment}
\end{table*}

To analyze the relevance of different contextual cues, we perform an ablation study and report the results in 
Table~\ref{tab:res_featureAssessment}. We list our proposed In-ROI Sensitivity~(IRS), the Negative Log-Likelihood~(NLL), 
and the Average Displacement Error~(ADE) for four prediction horizons and eight variants of the CVAE model,
each using different input feature combinations. 
In order to keep the complexity of this analysis manageable, we treat the head and body orientation as one combined feature 
named head\&body by concatenating the orientations.
Along with the metrics itself Table~\ref{tab:res_featureAssessment} reports 50\% confidence intervals for all values in 
order to show the uncertainty of the metric estimation. This allows to better interpret the amount to which one model 
improves about another, but these intervals are not suited for concluding whether the models perform significantly different. 
For that, one actually has to look at, e.g.,~90\% confidence intervals for pairwise differences of metrics. We do not report 
such intervals due to space considerations but they have been computed and used to check the statistical significance of the 
statements made in the following. 

Based on our pairwise significance analysis, we additionally report in Table~\ref{tab:res_optimal_model} for each metric 
and prediction horizon the optimal CVAE model (input feature set). A model is deemed optimal if there is no other model with
a significantly better performance. In cases of ties, we report the model with the least amount of input features, i.e. the 
least complex model. 

\begin{table*}[ht!]
	\centering
	\caption{Optimal input feature set per metric and time horizon}
	\setlength{\tabcolsep}{5pt}
	\begin{tabular}{cc@{\hspace{10pt}}ccc@{\hspace{10pt}}ccc@{\hspace{10pt}}cc}
		\toprule
		& & IRS & & NLL & & ADE \\
		\toprule
		
		1\unit{s} & & motion+egodist+head\&body 	& & motion+map+egodist				& & motion+map+egodist \\
		2\unit{s} & & motion+map+egodist+head\&body	& & motion+map+egodist+head\&body	& & motion+map+egodist+head\&body \\
		3\unit{s} & & motion+map 					& & motion+map+egodist+head\&body	& & motion+map+egodist+head\&body \\
		4\unit{s} & & motion+map		 			& & motion+map+egodist+head\&body	& & motion+map+egodist \\
		
		\bottomrule
	\end{tabular}
	\label{tab:res_optimal_model}
\end{table*}

Using our proposed IRS metric, we first analyze 
the relevance of different contextual cues. In comparison to other ablation studies, this analysis thus focuses on 
the importance of different input feature combinations with respect to the overall system performance. By 
taking the specific requirements of the system function into account, one can ensure that the prediction 
model is perfectly tailored towards a downstream task (e.g.~AEB-P). As shown in Sec.~\ref{sec:exp_metric_assessment}, 
the results of general prediction metrics (e.g.~NLL) and the IRS metrics may vary. An evaluation 
based on a system-level metric can consequently avoid the use of unnecessary complex models.

It is obvious that the map is crucial for good prediction performance, when comparing models 1-4 
(without map) to models 5-8 (with map). The positive effect of using a map is particularly pronounced for 
prediction horizons of 2\unit{s} to 4\unit{s}. However, our further significance analysis indicates three deviations 
from this general trend. Enhancing model CVAE motion+egodist with map does not yield a significant performance 
improvement for 3\unit{s} and 4\unit{s} predictions. Additionally, no significant improvement for a prediction 
horizon of 1\unit{s} can be observed when comparing model 3 to 7 and 4 to 8.

Overall the IRS metric can be significantly increased from $96.1$\unit{\%} to $98.5$\unit{\%} 
for short prediction horizons~(1\unit{s}) and from $82.6$\unit{\%} to $93.3$\unit{\%} 
for long  prediction horizons~(4\unit{s}) by extending the conditioning of the CVAE motion model to all 
available contextual cues. The IRS of model CVAE motion+map+head\&body is with $93.4$\unit{\%} even a little
better, but this improvement is not significant.

As can be seen in Table~\ref{sec:exp_metric_assessment}, an additional feature does not 
always yield a significant performance improvement when using our system-level IRS metric based on our selection criterion, dataset, and model family. 
Short-term predictions (1\unit{s}, 2\unit{s}) benefit from considering egodist and head\&body features. For long-term predictions (2\unit{s}, 3\unit{s}, and 4\unit{s}), it is advantageous to use map features. Comparing to CVAE motion+map, we did not observe significant improvements for 3\unit{s}, and 4\unit{s} by adding egodist or head\&body. However, this could be due to limited head and body orientation labels in the dataset.

In a second step, we compare the IRS score of the models with the corresponding ADE and NLL scores. Based on a toy example, we have 
shown in Sec.~\ref{sec:exp_metric_assessment} that improvements in NLL do not necessarily pay off in the proposed 
application-specific IRS-metric. Indeed, we also make similar observations in the ablation study results of 
Table~\ref{tab:res_featureAssessment}. For example, model~8 significantly improves over model~7 for the 4\unit{s} prediction 
horizon in terms of NLL, while this is not the case in terms of IRS. The difference are also visible in Table~\ref{tab:res_optimal_model}
and especially for a prediction horizon of 3\unit{s} and 4\unit{s}. According to our selection criterion, NLL and ADE suggest to use all available input features, 
our IRS metric, on the other hand, indicates that features motion and map are sufficient.


\subsection{Qualitative Evaluation} \label{sec:qualitative-evaluation}

\begin{figure*}[ht!]
	\centering
	\includegraphics[trim=0 40 0 0,clip,width=\textwidth]{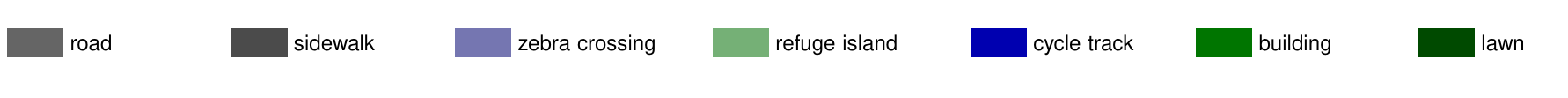}
	\begin{minipage}[t]{0.45\textwidth}
		\centering
		\subfloat[Zebra crossing (features: motion)]{\includegraphics[trim=0 0 0 0,clip,width=\textwidth]{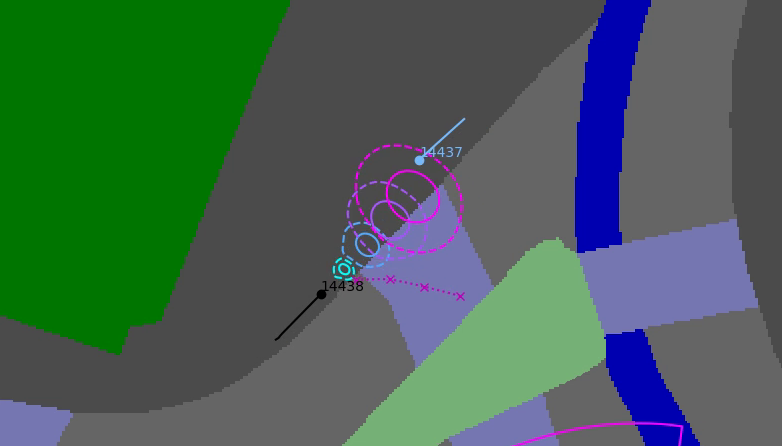}
			\label{fig:exp_qual_zebra_base}}	
		\vfil
		\subfloat[Zebra crossing (features: motion, map)]{\includegraphics[trim=0 0 0 0,clip,width=\textwidth]{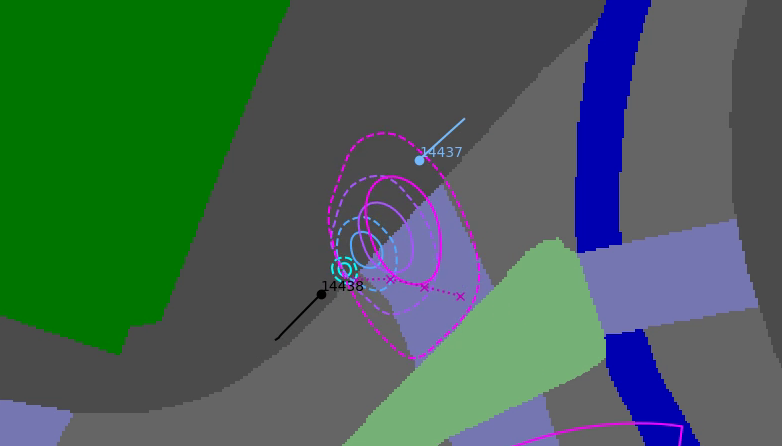}
			\label{fig:exp_qual_zebra_basemap}}
	\end{minipage}
	\hfill
	\begin{minipage}[t]{0.45\textwidth}
		\centering
		\subfloat[Crossing without zebra (features: motion)]{\includegraphics[trim=0 0 0 0,clip,width=\textwidth]{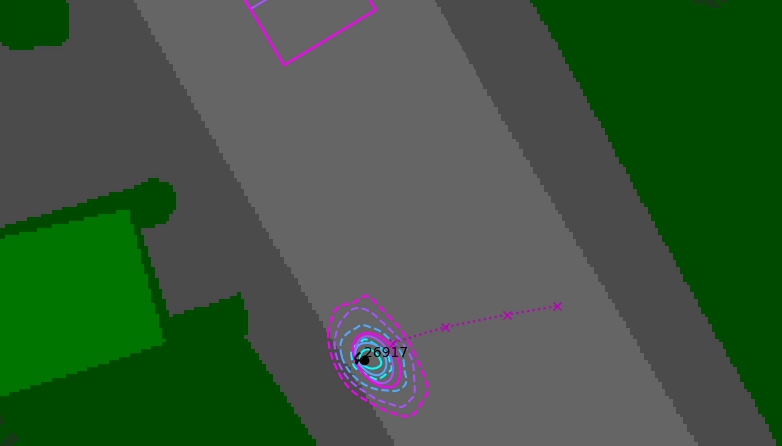}
			\label{fig:exp_qual_cross_base}}	
		\vfil
		\subfloat[Crossing without zebra (features: motion, map, egodist)]{\includegraphics[trim=0 0 0 0,clip,width=\textwidth]{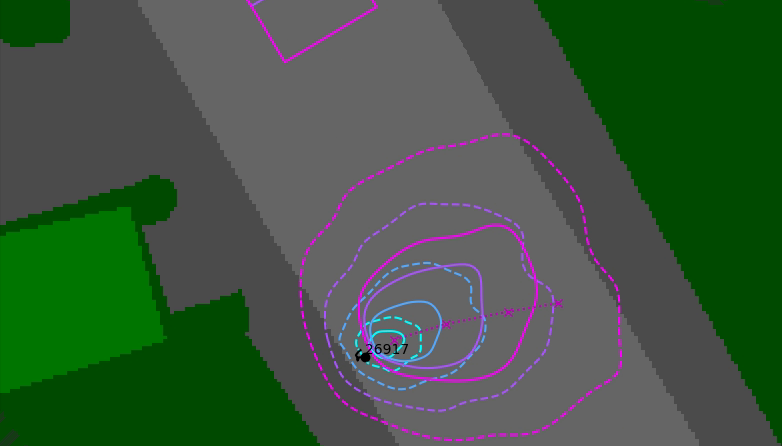}
			\label{fig:exp_qual_cross_basemapego}}
	\end{minipage}
	\caption{Pedestrian behavior prediction of the proposed model
		in two different scenarios. The circles indicate predictions
		for 1s (turquoise), 2s (blue), 3s (purple), and 4s
		(magenta), while the dashed line with crosses indicate the
		ground truth future trajectory. In the first scenario,
		depicted in (a) and (b), a pedestrian is walking towards a
		zebra crossing after walking straight on the sidewalk, but
		parallel to the street. In the second scenario, depicted in
		(c) and (d), a pedestrian is starting to cross the street
		without a zebra crossing, after a car has reduced its speed
		(prediction of the car is illustrated by a magenta box at
		the top).}
	\label{fig:exp_qual}
\end{figure*}

The system-level feature relevance assessment in
Section~\ref{sec:exp_feature_relevance} indicates that CVAE-based
models with additional features can significantly outperform a
simple CVAE motion baseline model. In the following, we provide qualitative 
examples, which highlight the influence of features on pedestrian 
prediction. Fig.~\ref{fig:exp_qual} illustrates two
scenes that often occur in urban scenarios. In the first scene,
Fig.~\ref{fig:exp_qual_zebra_base}
and~\ref{fig:exp_qual_zebra_basemap}, a pedestrian is walking on the
sidewalk (dark gray) parallel to the street (light gray) and
approaching a zebra crossing (violet). The CVAE motion model in
Fig.~\ref{fig:exp_qual_zebra_base} does not have information about
the static environment and thus mainly predicts straight walking with
some uncertainty. Instead, the CVAE with map feature in
Fig.~\ref{fig:exp_qual_zebra_basemap} was able to learn that zebra
crossings increase likelihoods of pedestrians to cross and it
correctly skews the distribution towards the zebra crossing, while
still keeping a mode for straight walking. The second scene,
Fig.~\ref{fig:exp_qual_cross_base}
and~\ref{fig:exp_qual_cross_basemapego}, shows a pedestrian that stands
at the side of a street and waits for crossing, while a car is
decelerating to let the pedestrian pass. The CVAE motion model in
Fig.~\ref{fig:exp_qual_cross_base} with only pedestrian features is
not able to predict that the likelihood increases for the pedestrian
to start walking, while the CVAE model with map and ego vehicle
features in Fig.~\ref{fig:exp_qual_cross_basemapego} picks up this
information very fast and predicts the pedestrian to cross the street.


\section{Conclusion}
\label{sec:discussion}

With the shift from advanced driver assistance systems towards fully automated driving, novel requirements on pedestrian behavior prediction arise. In this paper, we argued that these requirements are not fully taken into account by established evaluation procedures -- particularly in terms of metrics and datasets that are usually used to quantitatively assess prediction performance. We proposed a system-level approach to bridge this gap: based on a large dataset comprising thousands of pedestrian-vehicle interactions, we analyzed human driving behavior, derived appropriate reaction patterns of an AD system, and finally specified corresponding requirements on a pedestrian behavior prediction component. Moreover, we proposed a novel evaluation metric that measures the fulfillment of these requirements. It eases interpretation of prediction performance from a system-level perspective and thus allows for balancing model complexity vs. system-level performance. Our contribution thus shall stimulate future research on system-level evaluation and optimization of prediction models.

The proposed metric was evaluated on a large-scale dataset comprising thousands of real-world pedestrian-vehicle interactions using a CVAE-based model. A thorough ablation study shed light on the relative importance of different features. We demonstrated that considering additional contextual cues does not always yield a significant performance improvement when using a system-level metric (i.e. our IRS metric). We also showed that results of general prediction metrics (e.g.~NLL) and system-level metrics differ. Consequently, an evaluation based on a system-level metric can avoid the use of unnecessary complex models, highlighting the importance of a system-level approach to pedestrian behavior prediction.

Future work could extend the ablation study towards additional datasets and contextual cues such as considering the influence of other traffic participants besides the ego vehicle (e.g. other pedestrians or vehicles). Furthermore, the evaluation of additional model types could yield interesting insights, such as analyzing models that make joint predictions for multiple traffic participants at once. This would allow comparing the IRS metric to multi-agent prediction metrics.
Finally, our future work will focus on the integration and optimization of the developed prediction component in an AD system.


\section{Acknowledgments}

This work is a result of the research project @CITY-AF –-- Automated Cars and Intelligent Traffic in the City: Automated Driving Functions. The project is supported by the Federal Ministry for Economic Affairs and Energy (BMWi), based on a decision taken by the German Bundestag. The authors are solely responsible for the content of this publication.

\appendices
\section{Implementation Details}
\label{sec:implementationDetails}

\subsection{Model Architecture}
\label{subsec:modelArchitecture}
\textbf{Recurrent encoder:} We compute an embedding of the observed pedestrian trajectory 
$x_{t-H:t}$ using a recurrent encoder. This encoder consists of two stacked Long Short-Term 
Memory (LSTM) cells producing a 128 dimensional embedding vector. Both cells use the same 
state size which is determined for each input feature combination by means of a  
hyperparameter search. The observation horizon $H$ is set to 10 time steps, which corresponds 
to one second. 

\textbf{Map encoder:} The static environment, represented as a bird's-eye view semantic grid, 
is encoded via a small Convolutional Neural Network (CNN). The CNN processes grids of size 
256$\times$256 pixels, containing the agent's local environment of size 25.6\,m$\times$25.6\,m 
centered around the position of the pedestrian at the last conditioning time step. The 
architecture of the CNN is defined by the shortcut notation:
\begin{equation*}
	C'_4\text{-}P\text{-}C'_8\text{-}P\text{-}C'_{16}\text{-}C'_{16}\text{-}P\text{-}
	C'_{32}\text{-}C'_{32}\text{-}P\text{-}C'_{64}\text{-}C'_{64}\text{-}P\text{-}
	F'_{512}\text{-}F_{100},
\end{equation*} 
where $C_i$ is a convolutional layer with $i$ filters, a stride of one and a filter size of 
3$\times$3, $P$ a max-pooling layer with non-overlapping 2$\times$2 regions, and $F_{i}$ a 
fully connected layer with~$i$ output features. A prime marks layers which apply a Rectified Linear Unit (ReLU) 
nonlinearity. 

\textbf{Feature transformers:} The predictive model and the inference model each use a Multilayer 
Perceptron (MLP) to transform feature vectors to parameters of an $n$-dimensional Gaussian distribution. 
The dimensionality $n$ is set to eight for the predictive model and to ten for the inference model. 
Both MLPs consist of three fully connected layers and utilize ReLU nonlinearities. The number 
of output features is identical in each layer and derived using a hyperparameter search. Missing features are replaced by a constant value of zero.

\subsection{Model training}
We perform a grid search to determine the best model for each input feature combination. 
The parameters of our hyperparameter search space are listed in 
Table~\ref{tab:hyperparameters}. In total, we train 60 models for each input feature 
combination and pick the best model based on the validation IRS. Models are trained 
for 3000 epochs using the Adam optimizer~\cite{KingmaB2015} with a constant learning rate of 0.001. To augment the 
training data, we randomly rotate the trajectories and environment maps. We use a batch size of 1024 for the models using the semantic 
map as an input feature and a batch size of 2048 for models without map. All weights of the model are reparametrized using weight 
normalization~\cite{SalimansKingma2016}.
\begin{table}[h!]
	\centering
	\caption{Hyperparameter search space}
	\begin{tabular}{ c | c | c }
		& Models with map 	& Models without map \\
		\hline
		State size of LSTM cells 		& 256, 384 			& 256, 384 \\
		Features per MLP layer  		& 256, 384, 512 	& 384, 512, 640 \\
		Random seeds  					& 1, \dots, 10 		& 1, \dots, 10 \\		
	\end{tabular}
	\label{tab:hyperparameters}
\end{table}

%
%
%
%
%

\ifCLASSOPTIONcaptionsoff
  \newpage
\fi



%



\bibliographystyle{./bibtex/IEEEtran} 
\bibliography{./bibtex/IEEEabrv,./bibtex/pedBehaviorPrediction}

%

\begin{IEEEbiography}[{\includegraphics[width=1in,height=1.25in,clip,keepaspectratio]{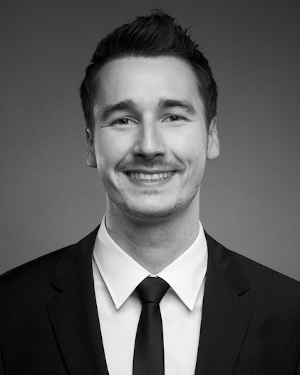}}]{Michael Herman}
	is a research scientist and sub-project lead at the Bosch Center for Artificial Intelligence working on machine learning-based motion prediction for automated driving. He received his Ph.D.~degree in computer science from the University of Freiburg in an industrial Ph.D. program, where his research focused on learning generalizable representations of an experts motivations from observed behavior in complex, unknown environments. His current research is focused on learning prediction models in multi-agent systems with a focus on automated driving use cases, while his research interests include Probabilistic Inference, Deep Learning, and Imitation Learning.
\end{IEEEbiography}

\begin{IEEEbiography}[{\includegraphics[width=1in]{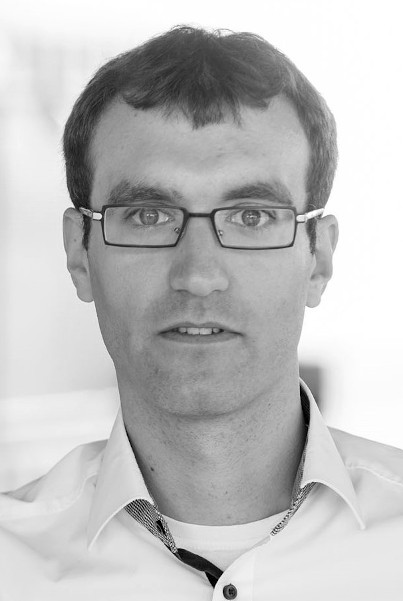}}]{J\"{o}rg Wagner}
received his M.Sc degree in electrical engineering and information technology from the Karlsruhe Institute of Technology, Germany, in 2014. He is a research engineer at Bosch Center for Artificial Intelligence (BCAI) working on machine learning-based motion prediction for automated driving. While working at BCAI, he is currently pursuing the Ph.D. degree with the University of Bonn, Germany.  His research interests include deep learning, interpretable and explainable~AI, computer vision and time series modeling.
\end{IEEEbiography}

\begin{IEEEbiography}[{\includegraphics[width=1in]{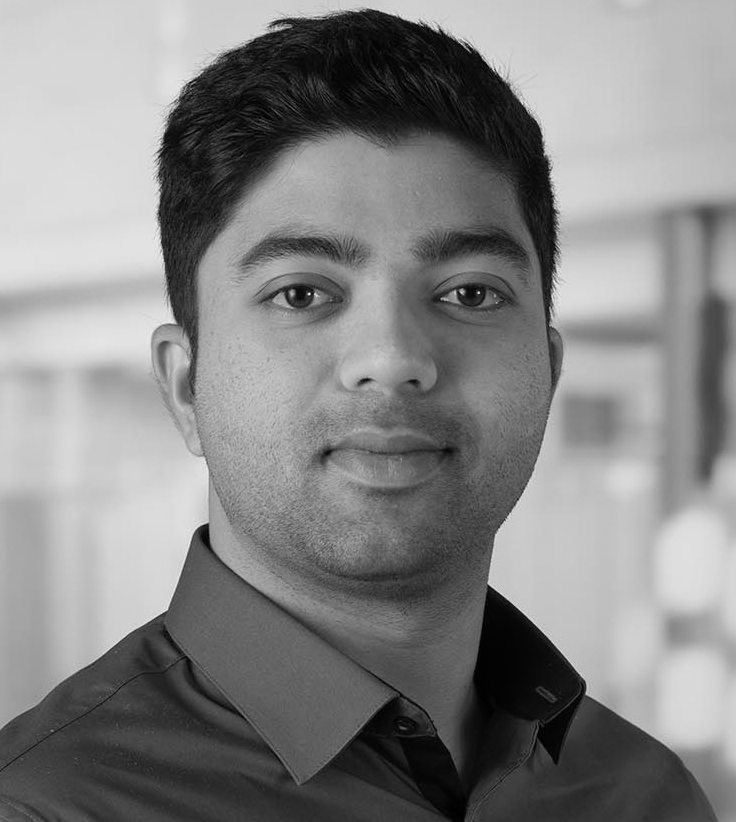}}]{Vishnu Prabhakaran}
received his M.Sc degree in Information Technology in 2018 from the University of Stuttgart, Germany, with a focus on machine learning and robotics. He is a research engineer at Bosch Center for Artificial Intelligence, working on multi-agent behavior prediction models for automated driving. His research interests include probabilistic inference, deep learning and time series modeling.
\end{IEEEbiography}

\begin{IEEEbiography}[{\includegraphics[width=1in,height=1.25in,clip,keepaspectratio]{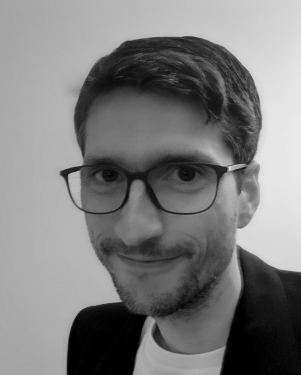}}]{Nicolas M\"{o}ser}
received his Dr. rer. nat. (Ph.D.) in 2011 from the University of Bonn, Germany, in experimental elementary particle physics at the ATLAS experiment at CERN, Geneva. In 2012 he joined Corporate Research of Robert Bosch GmbH, where he has been involved in various data mining and machine learning activities related to driver assistance and automated driving. 
\end{IEEEbiography}

\begin{IEEEbiography}[{\includegraphics[width=1in,height=1.25in,clip,keepaspectratio]{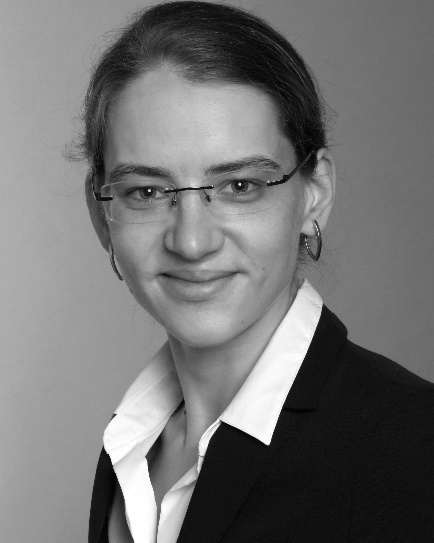}}]{Hanna Ziesche}
is a research scientist at the Bosch Center for Artificial Intelligence working on robotics and information theoretic deep reinforcement learning. She received her Dr. rer. nat. (Ph.D) in 2016 from the University of Karlsruhe, Germany, in theo\-re\-ti\-cal particle physics. In 2017 she joined Robert Bosch GmbH, where she has been involved in various projects on machine learning and reinforcement learning topics. 
\end{IEEEbiography}

\begin{IEEEbiography}[{\includegraphics[width=1in,height=1.25in,clip,keepaspectratio]{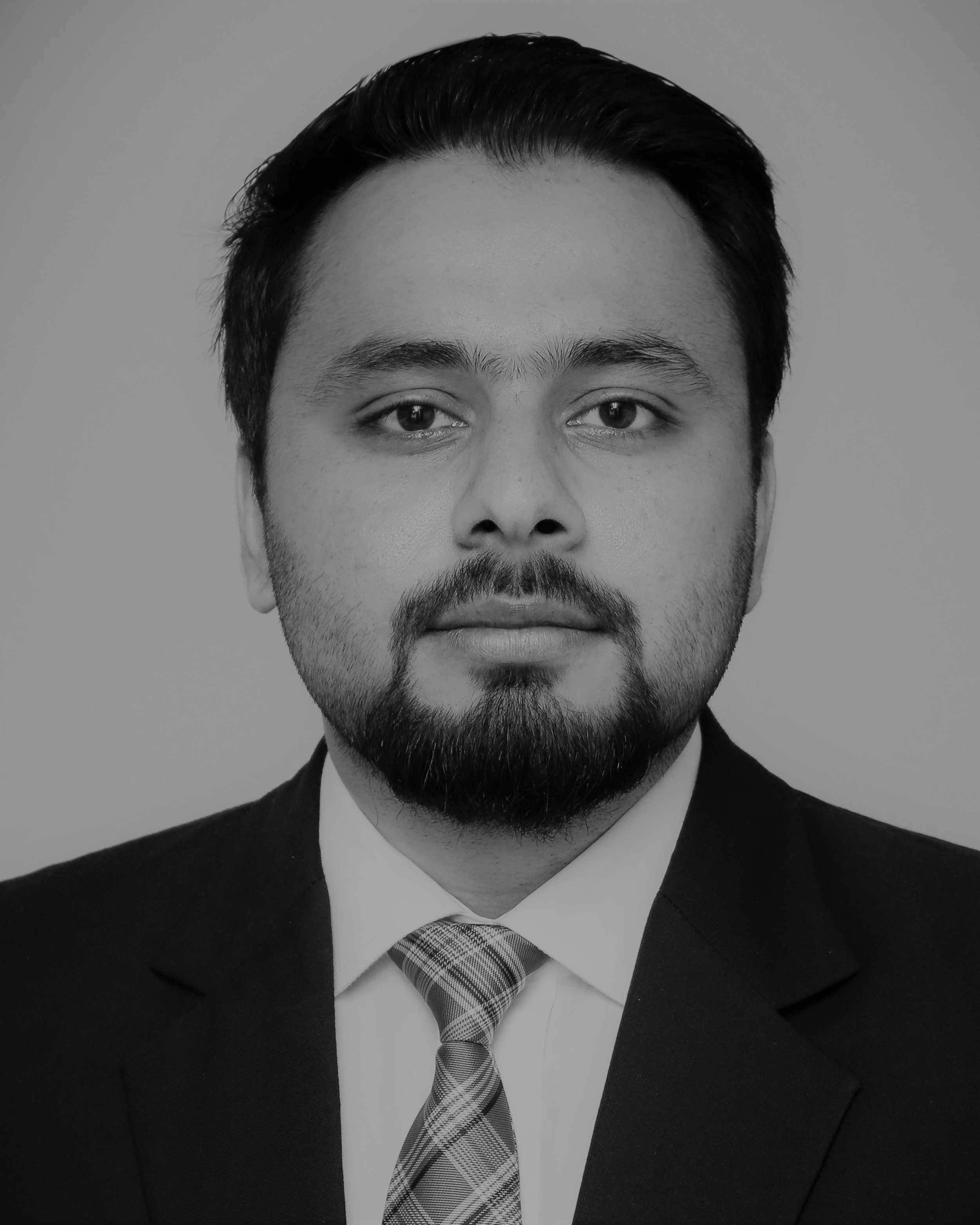}}]{Waleed Ahmed}
received his M.Sc degree in Automation and Robotics in 2018 from the Technical University of Dortmund, Germany, with a focus on robotics and machine learning. Since 2018, he is a development engineer at the Bosch Cognitive Systems Group with a focus on Automated Driving. He has been involved in various machine learning activities for automated driving and identification of driving strategies for driver assistance systems. 
\end{IEEEbiography}

\begin{IEEEbiography}[{\includegraphics[width=1in,height=1.25in,clip,keepaspectratio]{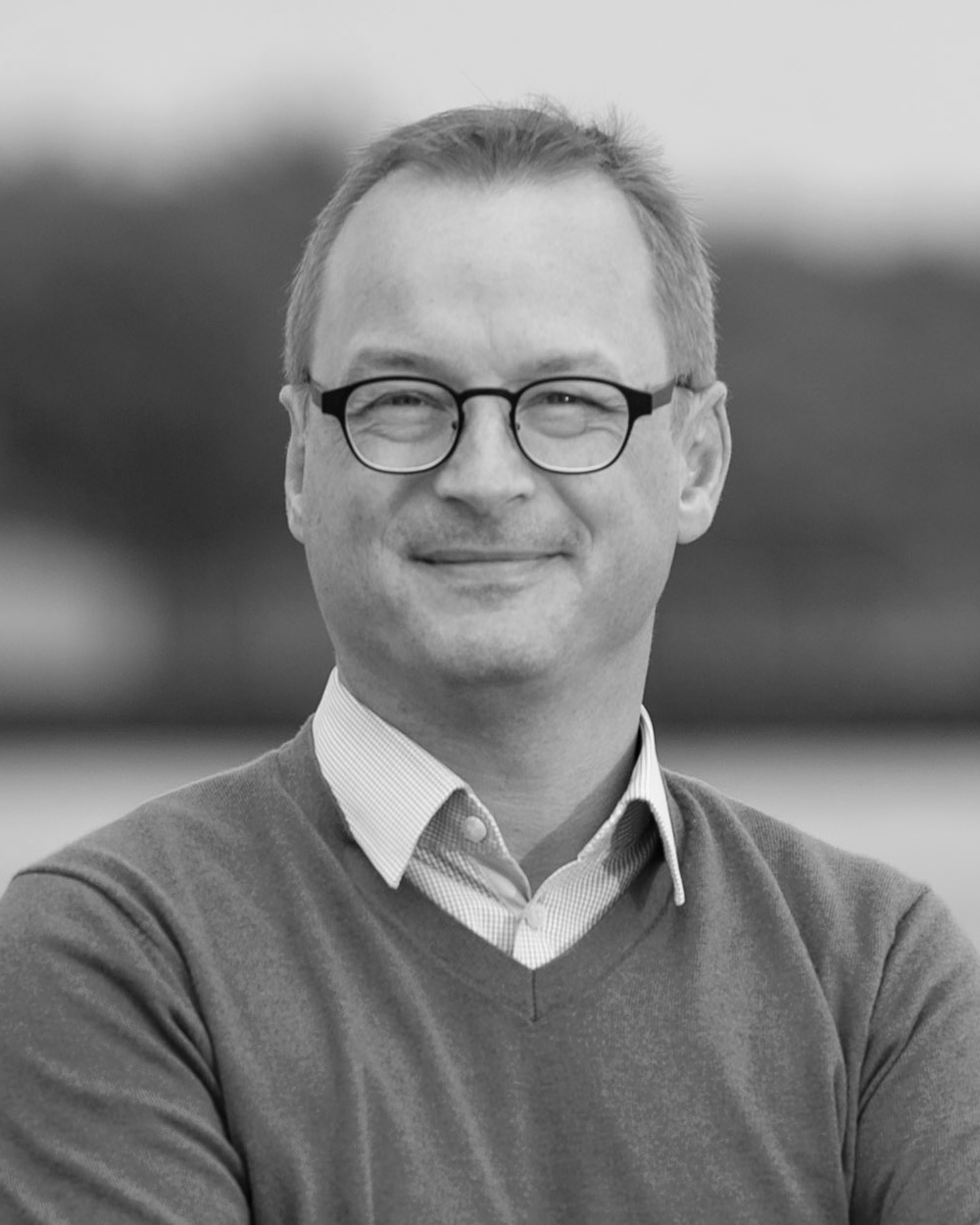}}]{Lutz B\"{u}rkle}
received his Dr. rer. nat. (Ph.D.) in physics from the University of Freiburg, Germany, in 2001. From 1997 he was a Research Scientist at the Fraunhofer Institute of Applied Solid State Physics in Freiburg, Germany. In 2002 he joined Robert Bosch GmbH, where he has been involved in the development of various driver assistant and automated driving systems. He is currently a project manager at Bosch Corporate Research in Renningen, Germany.
\end{IEEEbiography}

\begin{IEEEbiography}[{\includegraphics[width=1in]{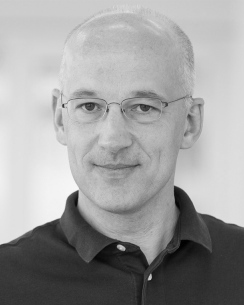}}]{Ernst Kloppenburg}
	received his Dr.-Ing. (Ph.D.) degree from University of Stuttgart in 1999 in Technical Cybernetics / Control Engineering. He is a senior expert at Bosch Center for Artificial Intelligence, working in the fields of probabilistic inference, and of verification of machine learning systems.  
\end{IEEEbiography}

\begin{IEEEbiography}[{\includegraphics[width=1in,height=1.25in,clip,keepaspectratio]{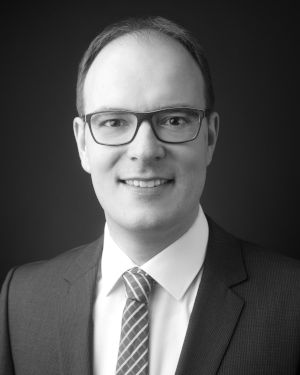}}]{Claudius Gl\"{a}ser}
	received his Dr.-Ing. (Ph.D.) degree in computer science from Bielefeld University, Germany, in 2012. From 2006 he was a Research Scientist with the Honda Research Institute Europe GmbH, Offenbach/Main, Germany, working in the fields of speech processing and language understanding for humanoid robots. In 2011, he joined the Corporate Research of Robert Bosch GmbH in Renningen, Germany, where he developed perception algorithms for driver assistance and highly automated driving functions. He is currently senior expert for sensor data fusion in autonomous systems. His research interests include environment perception, multimodal sensor data fusion, multi object tracking, and machine learning for highly automated driving.
\end{IEEEbiography}







\end{document}